\documentclass[lettersize,journal]{IEEEtran}  

\IEEEoverridecommandlockouts                              

\usepackage{cite}
\usepackage{amsmath,amssymb,amsfonts}
\usepackage{algorithmic}
\usepackage{graphicx}
\usepackage{textcomp}
\usepackage{setspace}
\usepackage{booktabs}
\usepackage{tikz}
\usepackage{hyperref}
\hypersetup{
    colorlinks,
    linkcolor={red!50!black},
    citecolor={blue!50!black},
    urlcolor={blue!80!black}
}
\usepackage{epsfig}
\usepackage{times} 
\usepackage{svg}
\usepackage[linesnumbered,ruled,vlined]{algorithm2e} 

\usepackage[nolist, nohyperlinks]{acronym}

\usepackage{xcolor}
\usepackage{bm}
\usepackage{breqn}
\usepackage{amsmath}
\usepackage{amssymb}
\usepackage{tabularx}
\usepackage{capt-of}

\usepackage{amsthm}

\newcommand{\reals}{\mathbb{R}}
\newcommand{\R}{\reals}
\newcommand{\Rnonneg}{\reals_{\geq 0}}
\newcommand{\Rplus}{\reals_{>0}}

\newcommand{\naturals}{\mathbb{N}}
\newcommand{\N}{\naturals}

\newcommand{\Acal}{\mathcal{A}}

\newcommand{\Ccal}{\mathcal{C}}
\newcommand{\Dcal}{\mathcal{D}}
\newcommand{\Ecal}{\mathcal{E}}
\newcommand{\Fcal}{\mathcal{F}}
\newcommand{\Gcal}{\mathcal{G}}
\newcommand{\Hcal}{\mathcal{H}}
\newcommand{\Ical}{\mathcal{I}}
\newcommand{\Jcal}{\mathcal{J}}

\newcommand{\Mcal}{\mathcal{M}}
\newcommand{\Ncal}{\mathcal{N}}
\newcommand{\Ocal}{\mathcal{O}}

\newcommand{\Rcal}{\mathcal{R}}

\newcommand{\Tcal}{\mathcal{T}}
\newcommand{\Ucal}{\mathcal{U}}
\newcommand{\Vcal}{\mathcal{V}}
\newcommand{\Wcal}{\mathcal{W}}
\newcommand{\Xcal}{\mathcal{X}}

\newcommand{\Zcal}{\mathcal{Z}}

\newcommand{\eqn}[1]{\begin{align} #1 \end{align}}
\newcommand{\eqnN}[1]{\begin{align*} #1 \end{align*}}

\newcommand{\bmat}[1]{\begin{bmatrix}#1\end{bmatrix}}

\newcommand{\norm}[1]{\left\Vert #1 \right \Vert}

\newcommand{\abs}[1]{\left | #1 \right |}

\newcommand{\argmin}[1]{\underset{\substack{#1}}{\text{argmin}}}

\theoremstyle{plain}
\newtheorem{theorem}{Theorem}

\newtheorem{lemma}{Lemma}
\newtheorem{problem}{Problem}
\newtheorem{definition}{Definition}
\newtheorem{prop}{Proposition}

\theoremstyle{definition}
\newtheorem{assumption}{Assumption}
\newtheorem{remark}{Remark}

\theoremstyle{remark}

\theoremstyle{definition}
\newtheorem*{runningexample}{Running Example}
\newtheorem*{runningexamplecont}{Running Example (cont.)}
\theoremstyle{remark}

\makeatletter
\let\NAT@parse\undefined
\makeatother
\usepackage{hyperref}
\hypersetup{
colorlinks, 
    linkcolor={red!50!black},
    citecolor={blue!50!black},
    urlcolor={blue!80!black}
}
\usepackage{cleveref}

\crefformat{problem}{Problem~#2#1#3}
\crefformat{assumption}{Assumption~#2#1#3}

\usepackage[nolist,nohyperlinks]{acronym}

\acrodef{CP}[CP]{conformal prediction}
\acrodef{MAICP}[MA-ICP]{multi-agent interactive conformal prediction}

\newcommand{\br}{\boldsymbol{r}}

\newenvironment{authorbiography}[2][]{%
  \par\noindent
  \begin{minipage}[t]{\columnwidth}
    \begin{minipage}[t]{1in}
      \vspace{0pt}
      #1
    \end{minipage}
    \hspace{0.8em}%
    \begin{minipage}[t]{\dimexpr\columnwidth-1.15in\relax}
      \vspace{0pt}
      \small
      \textbf{#2}\ 
}{%
    \end{minipage}
  \end{minipage}
  \par\vspace{0.8em}
}

\title{Distributed Interactive Conformal Prediction for Multi-Agent Systems}

\crefname{definition}{Def.}{Defs.}
\author{Daniel M. Cherenson and Dimitra Panagou
\thanks{DISTRIBUTION STATEMENT A. Approved for public release; distribution is unlimited. OPSEC11252}
\thanks{This work was supported by the Automotive Research Center (ARC), a US Army Center of Excellence for modeling and simulation of ground vehicles, under Cooperative Agreement W56HZV-24-2-0001 with the US Army DEVCOM Ground Vehicle Systems Center (GVSC).}
\thanks{Daniel Cherenson is with the Department of Robotics, University of Michigan, Ann Arbor, MI, USA
        {\tt dmrc@umich.edu} (Corresponding author)} %

\thanks{Dimitra Panagou is with the Department of Robotics and Department of Aerospace Engineering, University of Michigan, Ann Arbor, MI, USA {\tt dpanagou@umich.edu}}%
}

\begin{document}

\maketitle
\begin{abstract}
Multi-agent systems increasingly rely on learning-enabled components and distributed algorithms for perception, estimation, and control, introducing uncertainty that makes it challenging to satisfy mission requirements. Conformal prediction quantifies uncertainty with distribution-free guarantees, but existing methods do not address the challenging combination of dependence among robots, distributed calibration, and the distribution shift induced when calibrated uncertainty bounds modify the closed-loop system. We introduce Multi-Agent Interactive Conformal Prediction (MA-ICP), a distributed framework for computing robustness margins in heterogeneous systems to satisfy mission requirements with high probability. MA-ICP aggregates robot-level nonconformity scores, which measure errors in modeled or estimated quantities, into one mission-level score per robot class, allowing dependence among robots within each mission. Mission-level scores and quantiles are computed using neighbor-to-neighbor communication in a finite number of communication rounds. Updating the robustness margins changes robot trajectories and distributed algorithm outputs, so we use class-dependent score sensitivities to upper bound the shift in mission-level scores. We then establish calibration-conditional coverage guarantees per robot class for the updated margins. We demonstrate MA-ICP on cooperative payload control, formation control with obstacle avoidance, and distributed target tracking missions.
\href{https://github.com/dcherenson/maicp}{[Code]}\footnote{\scriptsize\url{https://github.com/dcherenson/maicp}}
\end{abstract}
\section{Introduction}

Collaboration in multi-agent systems enables robots to perform tasks that are difficult or inefficient for a single robot~\cite{rizk2019cooperative}, including distributed sensing and target tracking~\cite{robin2016multi}, information gathering~\cite{dutta2021multi}, and search and rescue~\cite{chung2018survey,queralta2020collaborative}. These systems increasingly rely on learning-enabled components for perception, dynamics modeling, prediction, estimation, and control, which can improve performance in complex and uncertain environments~\cite{zhang2021multi}. However, learned models, noisy data, imperfect dynamics, and approximate distributed computations introduce uncertainty into the quantities used to make control and planning decisions. In safety-critical applications, these errors can directly affect collision avoidance, connectivity, tracking, localization, and other mission requirements, motivating methods that quantify uncertainty and incorporate the resulting bounds into the closed-loop system~\cite{brunke2022safe, naveed2025enabling, wang2025uncertainty,cherenson2026distributionally}.

Uncertainty quantification is particularly challenging in cooperative multi-robot systems because the uncertainty experienced by one robot is coupled to the behavior of the rest of the team. Robots exchange measurements and algorithm states, react to one another, e.g., through their control policies, and execute distributed optimization and estimation algorithms whose outputs depend on information propagated through the communication network. As a result, the errors generated by robots in the same mission can be strongly dependent, while different robot classes, such as ground and aerial robots with heterogeneous sensing and dynamics models, may exhibit different error characteristics.

An additional difficulty arises when uncertainty bounds are incorporated into controllers or distributed algorithms. These bounds, which we refer to as robustness margins, are used to account for errors in modeled or estimated quantities. For example, a robust {control barrier function (CBF)} can use a robustness margin to bound the difference between a robot's predicted and actual acceleration. However, changing the margin also changes the control inputs and trajectories, potentially shifting the distribution of the data used to calibrate it. Thus, the robustness margins cannot generally be calibrated offline.



In this paper, we focus on \acf{CP}, a statistical method that constructs finite-sample, distribution-free uncertainty bounds from calibration data under an exchangeability assumption~\cite{vovk2005algorithmic,angelopoulos2023conformal}. The calibration data consist of nonconformity scores, which are scalar measures of the discrepancy between predicted or estimated quantities and their corresponding reference values, with larger scores indicating greater errors. Recently, \ac{CP} has been applied to robust motion planning, perception, and safety-critical control in autonomous systems~\cite{lindemann2023safe,marques2024quantifying,muthali2023multi,zhou2024safety,cherenson2026staggered,dixit2025perceive}. In our setting, however, the exchangeability assumption of the calibration and test data of split \ac{CP}~\cite{angelopoulos2023conformal} creates two challenges: robot trajectories in the same mission are dependent, and updating the calibrated robustness margins can change the distribution of the data collected in subsequent missions.

To address these challenges, we introduce \textbf{Multi-Agent Interactive Conformal Prediction (MA-ICP)}\acused{MAICP}, a distributed method for calibrating class-dependent robustness margins in heterogeneous multi-robot systems. We consider mission-level requirements that allow a prescribed number of robots in each class to violate their individual requirements. For example, in a target tracking mission with five robots, we may require that at least four robots maintain their tracking errors within the prescribed bounds throughout the mission. 
We capture this requirement by aggregating the robot-level nonconformity scores into one mission-level score per robot class. The collection of mission-level scores forms the calibration data, allowing dependence among robots participating in the same mission.

\ac{MAICP} uses these mission-level scores to iteratively calibrate the robustness margins. At each iteration, the robots execute a batch of missions under the current margins and compute the resulting mission-level scores and calibration quantiles through communication. The mission-level score and calibration quantile are both order statistics, i.e., values selected from a collection ordered from smallest to largest. The first uses robot-level scores within a class and mission, while the second uses mission-level scores across missions. Since changing the margins can alter the robot trajectories and distributed algorithm outputs, we account for the resulting distribution shift using class-dependent score sensitivity bounds. We also construct upper bounds on the order statistics after a finite number of distributed optimization iterations, without requiring asymptotic convergence. Combining these bounds allows us to establish calibration-conditional coverage guarantees for each robot class under the updated robustness margins. We demonstrate the method on cooperative payload force allocation, distributed formation control with obstacle avoidance, and distributed target tracking. In summary, our contributions are as follows:
\begin{itemize}
    \item We introduce mission-level nonconformity scores that allow a user-specified number of robot-level score violations while accounting for dependence among robots participating in the same mission.
    \item We develop a distributed method for computing mission-level scores and calibration quantiles through neighbor-to-neighbor communication, with valid upper bounds on the order statistics after a finite number of distributed optimization iterations.
    \item We derive sufficient conditions for bounding score sensitivity through coupled robot dynamics and distributed algorithms, and establish finite-sample, calibration-conditional coverage guarantees for each robot class under updated robustness margins.
\end{itemize}

\section{Related Work}
\label{sec:related_work}

\subsection{Conformal Prediction for Multi-Agent Systems}

Conformal prediction has been applied to planning and control in multi-agent environments to quantify uncertainty and provide probabilistic guarantees of safety. Several methods consider a robot operating among other agents whose actions cannot be directly controlled~\cite{lindemann2023safe,muthali2023multi,yu2026stl}. These approaches use \ac{CP} to construct prediction regions for the uncertain trajectories of surrounding agents, which are then incorporated into motion planning, reachability analysis, or temporal logic constraints. In these settings, the other agents are treated as uncontrollable agents rather than a collaborative team.

Other work has considered \ac{CP} in cooperative multi-robot systems. For example,~\cite{wang2024probabilistically} uses conformal prediction to provide probabilistic guarantees for language-based multi-robot task planning, while~\cite{huriot2025safe} incorporates uncertainty in neighboring agent predictions into decentralized control barrier function constraints. Although these methods use \ac{CP} for multi-robot planning and control, they do not address the iterative calibration of robustness margins that influence the distributed algorithms of a heterogeneous robot team.


\subsection{Interactive Conformal Prediction}
When uncertainty bounds are incorporated into control policies, updating the bounds can change the trajectories and errors observed during subsequent executions. This circular dependence has been addressed for interactive planning and control in~\cite{mirzaeedodangeh2026safe,mirzaeedodangeh2026robust}. These methods iteratively collect calibration data under the current policy and account for the distribution shift induced by subsequent policy updates using bounds on the change in nonconformity scores.

Related work has considered distribution shifts induced by interactions and changes in agent policies through off-policy conformal prediction~\cite{kuipers2024conformal} and iterative calibration~\cite{binny2025moved}. The former constructs joint trajectory prediction regions using density-ratio reweighting with marginal coverage guarantees, while the latter establishes convergence to a fixed-point distribution under a contraction condition rather than coverage after each policy update. AdaptNC~\cite{tumu2026adaptnc} adapts nonconformity scores and conformal thresholds online, but its coverage guarantees concern long-run average miscoverage. These methods do not directly establish finite-sample, calibration-conditional coverage for class-dependent mission requirements under updated robustness margins. In MA-ICP, we account for the distribution shift induced by these updates using score sensitivity bounds and compute the mission-level calibration quantities distributively.

\subsection{Federated and Distributed Conformal Prediction}


Federated conformal prediction methods allow uncertainty bounds to be calibrated using data held across multiple devices~\cite{lu2023federated,zhu2024federated,wen2026efficient}. While these methods commonly use a central server, distributed approaches compute calibration quantiles using communication among neighboring agents. In particular,~\cite{wen2025distributed} develops quantile-based and histogram-based distributed conformal prediction methods, while~\cite{wang2026distributeddata} considers distributed quantile estimation using binary search.

Although these methods allow calibration quantities to be computed without centralizing the complete calibration dataset, they do not directly account for the change in the score distribution caused by updating robustness margins in a coupled multi-robot system. In MA-ICP, we compute mission-level scores distributively and construct finite-iteration upper bounds on the order statistics, which are combined with the score sensitivity bounds in the robustness margin update.

\section{Preliminaries}
\label{sec:prelim}

For $n\in\N$, let $[n]:=\{1,\ldots,n\}$. For a collection of vectors $\{x_i\}_{i=1}^N$, $\operatorname{col}(x_1,\ldots,x_N)$ denotes column-wise concatenation. For an event $\mathcal E$, $\mathbf{1}_{\mathcal E}$ denotes its indicator. For vectors $v,w$ of equal dimension, $v\preceq w$ denotes component-wise inequality. For $a\in\R$, let $[a]_+:=\max\{0,a\}$ denote the positive part of $a$. For a positive definite matrix $D\in\R^{n\times n}$, we define the weighted Euclidean norm $\norm{x}_D:=\sqrt{x^\top D x}$. For scalars $x_1,\ldots,x_n$, let $x_{[1]}\leq\cdots\leq x_{[n]}$ denote the corresponding order statistics, where $x_{[k]}$ is the $k$th smallest value. 
We use $\mathbb P$ to denote probability under the joint law of the calibration and subsequent missions. For a $\sigma$-algebra $\Acal$, $\mathbb P(E\mid\Acal)$ denotes the conditional probability of an event $E$ given the information in $\Acal$. A random quantity is $\Acal$-measurable if its value is determined by this information.

\subsection{Conformal Prediction under Distribution Shift}
\label{sec:prelim_cp}
Conformal prediction constructs uncertainty bounds for a predictor using nonconformity scores that measure the discrepancy between predicted and observed quantities~\cite{vovk2005algorithmic,angelopoulos2023conformal}. Let $\Dcal:=(S^{(1)},\ldots,S^{(n)})$ denote calibration data consisting of $n\in\N$ nonconformity scores $S^{(1)},\ldots,S^{(n)}\in\R$, and let $S\in\R$ denote the score of a future observation. 
Standard conformal prediction provides marginal coverage guarantees when $S,S^{(1)},\ldots,S^{(n)}$ are exchangeable, meaning that their joint distribution is invariant under permutations. For the calibration-conditional guarantees considered in this paper, we make the stronger assumption that these scores are independent and identically distributed (i.i.d.).

Let $\alpha,\delta\in(0,1)$ denote the allowable miscoverage and calibration failure probabilities, respectively. We seek a threshold $q\in\R$ such that the probability of a future score exceeding $q$ is at most $\alpha$, with probability at least $1-\delta$ over the calibration data. To construct this threshold, let $B_n\sim\operatorname{Binomial}(n,1-\alpha)$ and define
\eqn{
k:=\min\left\{r\in[n]:\Pr(B_n\leq r-1)\geq1-\delta\right\}.
\label{eq:prelim_binomial_rank}
}
Provided $\delta\geq(1-\alpha)^n$, the rank $k$ is well defined. Choosing $q:=S_{[k]}$ and applying~\cite[Prop.~2b]{vovk2012conditional} we have
\eqn{
\Pr_{\Dcal}\left\{\Pr\left(S\leq q\mid\Dcal\right)\geq1-\alpha\right\}\geq1-\delta.
\label{eq:prelim_conditional_cp}
}
Here, the inner probability is taken over the future score conditional on the calibration data, while the outer probability is taken over the calibration data.

The guarantee in~\eqref{eq:prelim_conditional_cp} might become invalid when the robustness margins are updated, since the resulting control policies and distributed algorithms may change the distribution of the future score $S$. To account for this distribution shift, let $\widetilde S$ denote the score under the updated margins, coupled with $S$ using the same underlying randomness. Suppose there exists a function $M:\R^n\to\Rnonneg$ such that $\widetilde S\leq S+M(\Dcal)$ almost surely. Since $S\leq q$ implies $\widetilde S\leq q+M(\Dcal)$, we obtain
\eqn{
\Pr_{\Dcal}\left\{\Pr\left(\widetilde S\leq q+M(\Dcal)\mid\Dcal\right)\geq1-\alpha\right\}\geq1-\delta.
\label{eq:prelim_arcp}
}
This follows the robustification principle of adversarially robust conformal prediction~\cite{gendler2022adversarial}.

Finally, the mission-level scores and calibration quantiles required by MA-ICP are both order statistics, which can be computed using the pinball loss~\cite{koenker1978regression}, defined as $\ell_\tau(v):=\tau[v]_++(1-\tau)[-v]_+$ for $\tau\in(0,1)$. For any collection of scalars $a_1,\ldots,a_n$, choosing $\tau_k:=(k-1/2)/n$ gives $a_{[k]}$ as the unique minimizer of $\sum_{m=1}^n\ell_{\tau_k}(a_m-q)$ over $q\in\R$, even when values are repeated, as shown in~\Cref{sec:distributed_order_statistic}.

\section{Problem Formulation}
\label{sec:problem}
{Consider a heterogeneous multi-agent system with robot set $\Vcal:=\{1,\ldots,N\}$ and mission horizon $[0,T]$, where $T\in\Rplus$.} Each robot $i\in\Vcal$ belongs to a class $c \in \Ccal = \{1,\ldots,N_{\mathrm{cls}}\}$. Each robot is assigned to a class via the mapping $C : \Vcal \to \Ccal$. The set of all robots belonging to class $c$ is $    \Vcal_c
    :=
    \left\{
        i\in\Vcal:
        C(i)=c
    \right\}.$ 
Robots within each class share the same dynamics structure and use the same sensors, control policies, and distributed algorithms, but may have different physical parameters. 
The state and control input of robot $i$ are $x_i\in\Xcal_{C(i)}\subseteq\R^{n_{C(i)}}$ and $u_i\in\Ucal_{C(i)}\subseteq\R^{m_{C(i)}}$, respectively. Let $x:=\operatorname{col}(x_1,\ldots,x_N)\in\Xcal$, where $\Xcal:=\prod_{i\in\Vcal}\Xcal_{C(i)}$, denote the joint robot state. The robot dynamics are
\eqn{\dot{x}_i=f_{C(i)}(x,u_i,w_i;\theta_i),\label{eq:robot_dynamics}}
where $\theta_i\in\Theta_{C(i)}\subseteq\R^{n_{\theta,C(i)}}$ denotes the physical parameters of robot $i$, $w_i\in\Wcal_{C(i)}\subseteq\R^{n_{w,C(i)}}$ denotes exogenous inputs, and $f_c:\Xcal\times\Ucal_c\times\Wcal_c\times\Theta_c\to\R^{n_c}$ is Lipschitz in state and control for each fixed $w\in\Wcal_c$ and $\theta\in\Theta_c$. {Note that $f_{C(i)}$ depends on the joint robot state $x$, allowing the robot dynamics to be coupled.}

Let
$x^{\mathrm{env}}\in\Xcal^{\mathrm{env}}$ collect
application-dependent states that are not part of the robot joint state,
e.g., targets, obstacles, or landmarks. Robot $i$
receives a local measurement using the class-dependent measurement model:
\eqn{o_i=h_{C(i)}(i,x,x^{\mathrm{env}},\nu_i),\label{eq:local_measurement}}
where
$h_c:\Vcal_c\times\Xcal\times\Xcal^{\mathrm{env}}\times\R^{n_{\nu,c}}\to\Ocal_c$
and $\nu_i\in\R^{n_{\nu,C(i)}}$ denotes measurement noise, for every $c \in \Ccal$. The measurement
$o_i$ may contain information about the robot, neighboring robots,
targets, or other application-dependent quantities.

\subsection{Communication Model and Distributed Algorithms}
The robots interact over a communication graph $\Gcal(x)=(\Vcal,\Ecal(x))$, which may be fixed or state dependent. {The neighbor set of robot $i$ is $\Ncal_i(x):=\{k\in\Vcal\setminus\{i\}:(i,k)\in\Ecal(x)\}$.}

The robots execute distributed algorithms using local measurements and messages received from neighboring robots. Each class $c\in\Ccal$ is associated with a scalar robustness margin
$
    r_c
    \in
    \Rcal_c
    \subseteq
    \Rnonneg,$
and the vector of all class margins is
\eqn{
    \br
    :=
    \operatorname{col}
    \left(
        (r_c)_{c\in\Ccal}
    \right)
    \in
    \Rcal,
    \qquad
    \Rcal
    :=
    \prod_{c\in\Ccal}
    \Rcal_c.
    \label{eq:joint_margins}
}

Let $\xi_i\in\Xi_{C(i)}\subseteq\R^{n_{\xi,C(i)}}$ denote the internal state
maintained by robot $i$. Robot $i$ sends a message
$m_{ik}\in\Mcal_{C(i)}$ to each $k\in\Ncal_i(x)$. We let
$\hat m_i\in\widehat\Mcal_i$ collect the most recently available
incoming messages, where
$
\widehat\Mcal_i
:=
\prod_{k\in\Vcal\setminus\{i\}}
\Mcal_{C(k)}.$ Let $z_i\in\Zcal_{C(i)}$ denote the output of the distributed algorithm maintained by robot $i$.
Using these variables, we model distributed algorithms as follows:
\eqn{
\begin{aligned}
m_{ik}
&=
M_{C(i)}
\left(
    \xi_i,
    o_i,
    r_{C(i)},
    k
\right),
\\
\xi_i^+
&=
\Phi_{C(i)}
\left(
    \xi_i^-,
    o_i,
    \hat m_i^-,
    r_{C(i)}
\right),
\qquad
t\in\Tcal_i^{\mathrm{evt}},
\\
z_i
&=
H_{C(i)}
\left(
    \xi_i,
    o_i,
    \hat m_i,
    r_{C(i)}
\right),
\end{aligned}
\label{eq:distributed_algorithm}
}
where $\Tcal_i^{\mathrm{evt}}\subseteq[0,T]$ is the set of algorithm
update events for robot $i$,
$
M_{C(i)}:
\Xi_{C(i)}
\times
\Ocal_{C(i)}
\times
\Rcal_{C(i)}
\times
\Vcal
\to
\Mcal_{C(i)}
$
is the message map,
$
\Phi_{C(i)}:
\Xi_{C(i)}
\times
\Ocal_{C(i)}
\times
\widehat\Mcal_i
\times
\Rcal_{C(i)}
\to
\Xi_{C(i)}
$
is the discrete update map, and
$
H_{C(i)}:
\Xi_{C(i)}
\times
\Ocal_{C(i)}
\times
\widehat\Mcal_i
\times
\Rcal_{C(i)}
\to
\Zcal_{C(i)}
$
is the output map. The superscripts $-$ and $+$ are the algorithm states
immediately before and after an update event.

Each robot has a class-dependent policy
\eqn{
    u_i
    =
    \pi_{C(i)}
    \left(
        x_i,
        o_i,
        z_i;
        r_{C(i)}
    \right),
    \label{eq:agent_policy}
}
where $
    \pi_c:
    \Xcal_c
    \times
    \Ocal_c
    \times
    \Zcal_c
    \times
    \Rcal_c
    \to
    \Ucal_c$.
Although robot $i$ only applies the robustness margin of its own class $r_{C(i)}$, its trajectory and distributed algorithm output can depend on the entire margin vector $\br$ through coupled multi-agent dynamics.




\subsection{Missions and Nonconformity Scores}

Let $(\Omega,\Fcal_\Omega)$ denote the measurable space of mission realizations $\omega\in\Omega$, which specify the participating robots, initial robot and environment states, initial distributed algorithm states and stored messages, exogenous noise processes and event schedules, and any other random quantities required to determine the closed-loop execution. This statistical abstraction allows us to compare executions under different robustness margins using the same $\omega$, while keeping the underlying randomness fixed. The resulting trajectories, communication graphs, and state-dependent event times may differ between executions. Missions are thus random draws from $\Omega$.

\begin{definition}[Mission]
\label{def:mission}
For a mission realization $\omega\in\Omega$ and fixed robustness margin vector $\br\in\Rcal$, a \textbf{mission} is the closed-loop solution over $[0,T]$ generated by the robot dynamics~\eqref{eq:robot_dynamics}, measurement model~\eqref{eq:local_measurement}, communication graph $\Gcal(x)$, distributed algorithm~\eqref{eq:distributed_algorithm}, and control policy~\eqref{eq:agent_policy}. Given $(\omega,\br)$, this solution determines trajectories, measurements, interaction graph, and distributed algorithm states and outputs.
\end{definition}

{We omit the term ``nonconformity'' when referring to nonconformity scores.}
{For each robot $i\in\Vcal$ and mission $(\omega,\br)$, let $\emptyset\neq\Tcal_i^{\mathrm{score}}(\omega;\br)\subseteq[0,T]$ denote the times at which its score is evaluated. For continuous-time scores, this set is $[0,T]$, and for discrete algorithms, it may consist of the valid update times determined by the mission realization and the resulting closed-loop execution. Thus, the evaluation times may differ between robots and between executions under different robustness margins.}

\begin{definition}[Robot-Level Score]
\label{def:robot_score}
For each class $c\in\Ccal$, let $\rho_c:\Zcal_c\times\Zcal_c\to\Rnonneg$ denote an error measure. For robot $i\in\Vcal_c$, let $z_i^\star:[0,T]\times\Omega\times\Rcal\to\Zcal_c$ denote the reference quantity corresponding to the distributed algorithm output $z_i$. The \textbf{robot-level score} $S_i:\Omega\times\Rcal\to\Rnonneg\cup\{+\infty\}$ is
\eqn{
    S_i
    \left(
        \omega;
        \br
    \right)
    :=
    \sup_{t\in\Tcal_i^{\mathrm{score}}(\omega;\br)}
    \rho_{C(i)}
    \left(
        z_i
        \left(
            t;
            \omega,
            \br
        \right),
        z_i^\star
        \left(
            t;
            \omega,
            \br
        \right)
    \right).
    \label{eq:nonconformity_score}
}
\end{definition}

The score measures the (application-dependent) error associated with robot $i$ over one mission. We assume that the robot-level scores evaluated during calibration are finite almost surely. Realizations of~\eqref{eq:nonconformity_score} are used to construct the mission-level scores defined next. 


\begin{runningexample}[Target Tracking]
For distributed target tracking in~\Cref{sec:sim_tracking}, the quantities relevant to the score are $z_i=(\hat p_{i,t},P_{i,t})$ and $z_i^\star=(p_t^{\mathrm{tar}},P_{i,t})$, where $\hat p_{i,t}$ is the estimated target position, $p_t^{\mathrm{tar}}$ is the reference (true) target position, and $P_{i,t}\succ0$ is the corresponding covariance. We use the following score function,
\eqnN{
\rho_c(z_i,z_i^\star)
:=
\left[
\sqrt{
(\hat p_{i,t}-p_t^{\mathrm{tar}})^\top
P_{i,t}^{-1}
(\hat p_{i,t}-p_t^{\mathrm{tar}})
}
-q_{0,c}
\right]_+,
}
{where $q_{0,c}$ denotes the nominal ellipse radius. We use this score to calibrate uncertainty regions derived from the estimator covariance, which may not accurately represent target position uncertainty due to nonlinear measurements and non-Gaussian noise.}
{The score measures the largest excess beyond the nominal Mahalanobis radius over $\Tcal_i^{\mathrm{score}}(\omega;\br)$, and $S_i\leq r_c$ implies that the target lies within the corresponding uncertainty ellipse at every score evaluation time.}
\end{runningexample}

\begin{assumption}[Closed-loop well-posedness]
\label{ass:well_posed}
For almost every mission realization $\omega$, simultaneously for every $\br\in\Rcal$, the closed-loop multi-agent system defined by the robot dynamics, distributed algorithm~\eqref{eq:distributed_algorithm}, and policy~\eqref{eq:agent_policy} admits a unique solution on $[0,T]$.
\end{assumption}

The goal is to compute robustness margins $r_c$ that bound the robot-level scores with high probability. The application-dependent consequence of satisfying a robot-level requirement is specified separately for each case study.

\subsection{Mission-Level Score}

{For each class $c\in\Ccal$, let $N_c^{\mathrm{mis}}\in\{1,\ldots,\abs{\Vcal_c}\}$ denote the number of participating robots, and let $\Vcal_c^{\mathrm{mis}}:\Omega\to2^{\Vcal_c}$ specify which robots participate, with $\abs{\Vcal_c^{\mathrm{mis}}(\omega)}=N_c^{\mathrm{mis}}$. Together, these robots form the subteam $\Vcal^{\mathrm{mis}}(\omega):=\bigcup_{c\in\Ccal}\Vcal_c^{\mathrm{mis}}(\omega)$, where $\Vcal^{\mathrm{mis}}:\Omega\to2^{\Vcal}$. The subteam remains fixed throughout the mission and does not depend on the robustness margin $\br$, so subteams can execute missions in parallel with each mission contributing one calibration data point per class.}

\begin{definition}[Mission-Level Score]
\label{def:mission_score}
For each class $c\in\Ccal$, the user specifies a maximum of $\ell_c\in\{0,\ldots,N_c^{\mathrm{mis}}-1\}$ robot-level score violations per mission. We require at least $K_c:=N_c^{\mathrm{mis}}-\ell_c$ participating robot scores to satisfy their corresponding margins. Let $S_{c,[1]}(\omega;\br)\leq\cdots\leq S_{c,[N_c^{\mathrm{mis}}]}(\omega;\br)$ denote the order statistics of the indexed collection $(S_i(\omega;\br))_{i\in\Vcal_c^{\mathrm{mis}}(\omega)}$, counting repeated values separately. The \textbf{mission-level score} $S_c^{\mathrm{mis}}:\Omega\times\Rcal\to\Rnonneg\cup\{+\infty\}$ is
\eqn{S_c^{\mathrm{mis}}(\omega;\br):=S_{c,[K_c]}(\omega;\br).\label{eq:mission_score}}
\end{definition}



The condition $S_c^{\mathrm{mis}}(\omega;\br)\leq r_c$ requires at least $K_c$ participating class-$c$ robots to satisfy their score bounds, i.e., at most $\ell_c$ robots may violate them.

\begin{runningexamplecont}[Mission-Level Score]
Consider a target tracking mission with $N_c^{\mathrm{mis}}=5$ and $\ell_c=1$. Then $K_c=4$ and $S_c^{\mathrm{mis}}=S_{c,[4]}$, so $S_c^{\mathrm{mis}}\leq r_c$ requires at least four participating robots to satisfy their target position error bound over the score horizon. The robot whose score exceeds the bound need not be the same in different missions.
\end{runningexamplecont}

\subsection{Interactive Distribution Shift}

The robustness margins are inputs to the distributed algorithms and control policies in~\eqref{eq:distributed_algorithm} and~\eqref{eq:agent_policy}. Consequently, changing $\br$ changes the closed-loop solution in Definition~\ref{def:mission}, including the robot trajectories, measurements, interactions, and distributed algorithm outputs, and therefore changes the distribution of the resulting mission-level scores. Thus, mission-level scores collected under a fixed margin vector $\br$ characterize the distribution of $S_c^{\mathrm{mis}}(\omega;\br)$, which generally differs from the score distribution induced by a different margin vector.

To address this circular dependence, iterative \ac{CP} methods for interactive planning and control collect calibration data under a fixed policy and bound changes in the score induced by subsequent policy updates~\cite{mirzaeedodangeh2026safe,mirzaeedodangeh2026robust}. We will use the same principle to update the class-dependent robustness margins of the multi-agent system.

Accordingly, our method will proceed iteratively. At iteration $j\in\N$, the robots execute $M_j\in\N$ missions under the fixed margin vector
$
\br_j:=\operatorname{col}\left((r_{c,j})_{c\in\Ccal}\right)\in\Rcal
$
and use the resulting calibration data to compute $\br_{j+1}$. Along with the margins and the number of missions, the number of participating robots per class, score functions, and violation budgets are fixed before calibration. We let $\Hcal_j$ denote the information available at this point, represented by the pre-calibration $\sigma$-algebra, so these quantities are all $\Hcal_j$-measurable. Once the missions are completed, the collected data $\Dcal_j$ form the post-calibration $\sigma$-algebra $\Ical_j:=\sigma(\Hcal_j,\Dcal_j)$.


\begin{assumption}[Conditional Mission Sampling]
\label{ass:mission_sampling}
Let $\omega_j^{(m)}\in\Omega$, $m\in\{1,\ldots,M_j\}$, denote the realization of mission $m$, and let $\omega_j^+\in\Omega$ denote a subsequent mission realization. {Conditional on $\Hcal_j$, the mission-level score vectors $(S_c^{\mathrm{mis}}(\omega_j^{(m)};\br_j))_{c\in\Ccal}$ for $m\in\{1,\ldots,M_j\}$ and $(S_c^{\mathrm{mis}}(\omega_j^+;\br_j))_{c\in\Ccal}$ are i.i.d. In addition, $\omega_j^+$ is conditionally independent of the calibration mission realizations.}
\end{assumption}

For each robot $i$ participating in mission $m$, the realized robot-level score is $S_{i,j}^{(m)}
    :=
    S_i
    \left(
        \omega_j^{(m)};
        \br_j
    \right)$.
For each class $c\in\Ccal$, the corresponding mission-level scores are
\eqn{
    S_{c,j}^{\mathrm{mis},(m)}
    :=
    S_c^{\mathrm{mis}}
    \left(
        \omega_j^{(m)};
        \br_j
    \right),
    \qquad
    m\in\{1,\ldots,M_j\}.
    \label{eq:calibration_mission_scores}
}
By Assumption~\ref{ass:mission_sampling}, these mission-level scores are conditionally i.i.d. given $\Hcal_j$. The effective calibration sample size is therefore $M_j$, rather than the total number of robot-level scores generated within the missions.

To transfer the calibration result from $\br_j$ to a candidate updated margin vector, we next bound the change in the mission-level score as a function of the change in the robustness margins. Let $D\in\R^{N_{\mathrm{cls}}\times N_{\mathrm{cls}}}$ be a fixed positive definite matrix and define the margin distance $\Delta_j:\Rcal\to\Rnonneg$ by $\Delta_j(\br):=\norm{\br-\br_j}_D$.

Let $\Rcal_j\subseteq\Rcal$ be an $\Ical_j$-measurable, nonempty, closed, convex set of candidate robustness margins. One choice is a box of admissible margins, e.g., $\Rcal_j=\prod_{c\in\Ccal}[r_c^{\min},r_c^{\max}]$, where $0\leq r_c^{\min}<r_c^{\max}$ are known minimum and maximum robustness margins. We compare executions under $\br_j$ and $\br\in\Rcal_j$ using the same mission realization, so the initial conditions, noise realizations, and participating robots are unchanged, while the resulting trajectories, interaction graphs, and distributed algorithm outputs may differ.

\begin{definition}[Score Sensitivity Bound]
\label{def:score_sensitivity_bound}
For class $c\in\Ccal$ and iteration $j$, an $\Ical_j$-measurable coefficient $\kappa_{c,j}\in\Rnonneg$ is a \textbf{score sensitivity coefficient} over $\Rcal_j$ if, conditional on $\Ical_j$,
\eqn{S_c^{\mathrm{mis}}(\omega_j^+;\br)\leq S_c^{\mathrm{mis}}(\omega_j^+;\br_j)+\kappa_{c,j}\Delta_j(\br),\label{eq:mission_score_shift}}
holds for almost every mission $\omega_j^+$, simultaneously for every $\br\in\Rcal_j$.
\end{definition}

The bound applies to the resulting closed-loop executions, including changes in robot trajectories, communication graphs, and distributed algorithm update times induced by the robustness margins.

\begin{assumption}[Score Sensitivity]
\label{ass:score_sensitivity}
For every iteration $j\in\N$ and class $c\in\Ccal$, a finite $\Ical_j$-measurable coefficient $\kappa_{c,j}\in\Rnonneg$ satisfying Definition~\ref{def:score_sensitivity_bound} is available before computing $\br_{j+1}$.
\end{assumption}


The score sensitivity coefficient depends on the distributed algorithm and the quantity being calibrated. In~\Cref{sec:theory}, we give sufficient conditions for constructing $\kappa_{c,j}$ and then use these conditions to derive the coefficients for the cooperative payload force allocation case study in~\Cref{sec:sim_payload}.

\subsection{Problem Statement}

{Within each subteam, the participating robots compute their mission-level scores using a connected communication graph. Once the missions are complete, the robots use the update graph $\Gcal_j^{\mathrm{up}}=(\Vcal_j^{\mathrm{up}},\Ecal_j^{\mathrm{up}})$ to compute the calibration quantiles and update the robustness margins, where $\Vcal_j^{\mathrm{up}}\subseteq\Vcal$ includes all robots participating in the calibration missions. Both graphs are fixed, undirected, and connected during their respective distributed computations.} For each class $c\in\Ccal$, let $\alpha_c\in(0,1)$ denote the desired mission-level miscoverage and let $\delta_{c,j}^{\mathrm{cal}}\in(0,1)$ denote the allowable {calibration failure probability}.
\begin{figure*}
    \centering
    \includegraphics[width=\linewidth]{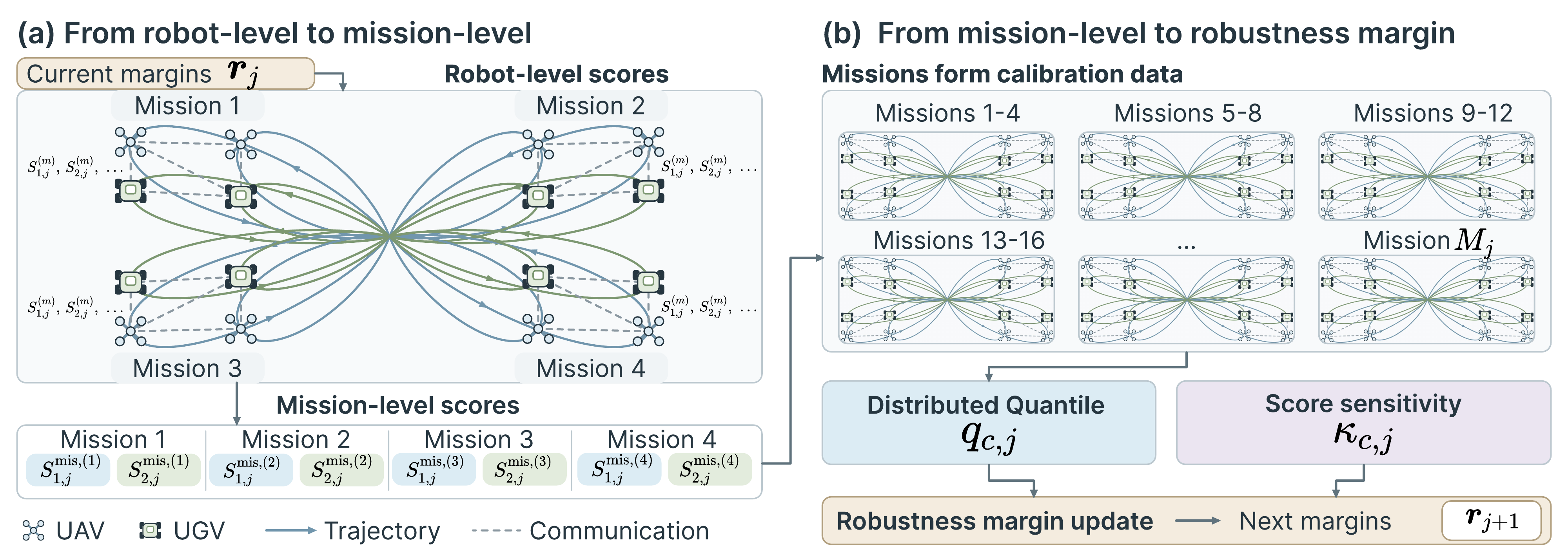}
    \caption{Overview of the proposed \ac{MAICP} calibration hierarchy. (a) Under the current robustness margins, heterogeneous teams execute their missions, potentially in parallel with subteams. Robot-level scores are computed from the resulting trajectories and aggregated within each class to obtain one mission-level score per class and mission. (b) Mission-level scores are used to compute the class-dependent quantiles, which are combined with the score sensitivities to obtain the robustness margins for the next iteration.}
    \label{fig:hero}
    \vspace{-10pt}
\end{figure*}
\begin{problem}[Multi-Agent Interactive Conformal Prediction]
\label{prob:maicp}
At iteration $j$, use the locally held calibration data and communicate over $\Gcal_j^{\mathrm{up}}$ to select a robustness margin vector $\br_{j+1}\in\Rcal_j$ such that the following holds almost surely for every class $c\in\Ccal$:
\eqn{
\begin{aligned}
&\mathbb P
\Bigg[
    \mathbb P
    \left(
        S_c^{\mathrm{mis}}
        \left(
            \omega_j^+;
            \br_{j+1}
        \right)
        \leq
        r_{c,j+1}
        \,\middle|\,
        \Ical_j
    \right)
    \geq
    1-\alpha_c
    \,\Bigg|\,
    \Hcal_j
\Bigg]
\\
&\hspace{3em}\geq
1-\delta_{c,j}^{\mathrm{cal}}.
\end{aligned}
\label{eq:desired_coverage}
}
\end{problem}

The inner probability represents mission-level coverage conditional on the completed calibration data, while the outer probability requires this coverage to hold with confidence at least $1-\delta_{c,j}^{\mathrm{cal}}$ given the information available before calibration.

\section{Methodology}
\label{sec:method}

\begin{figure}[t]
    \centering
    \includegraphics[width=\linewidth]{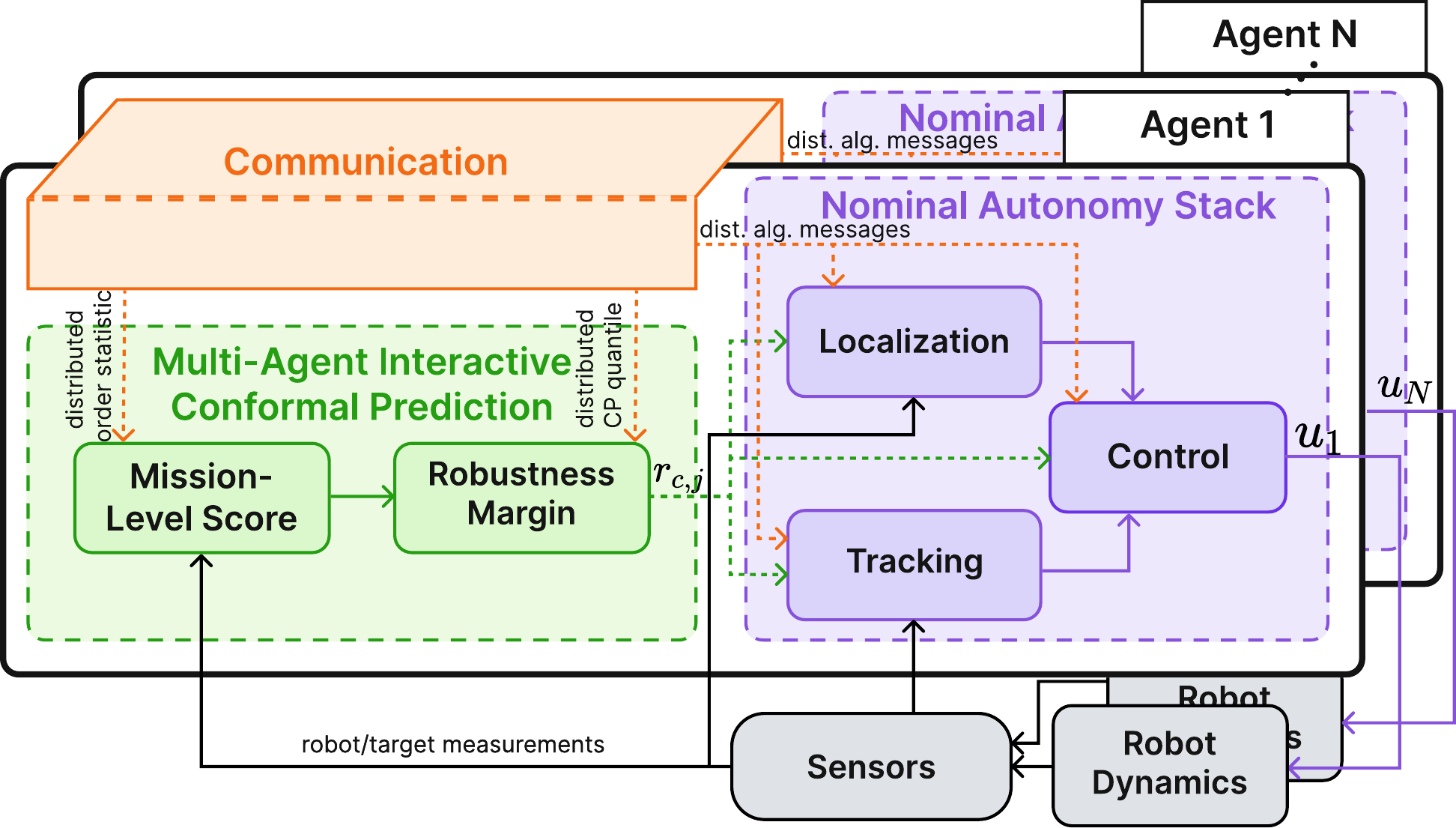}
    \caption{Block diagram highlighting how \ac{MAICP} fits into the autonomy stack.}
    \label{fig:block_diag}
\end{figure}

In this section, we introduce \textbf{\acf{MAICP}} to iteratively calibrate robustness margins while accounting for their effect on the closed-loop multi-agent system. 
Inspired by the single-agent iterative approach in~\cite{mirzaeedodangeh2026safe}, the robots execute $M_j$ missions under the fixed robustness margin vector $\br_j$ at iteration $j$. After each mission, the participating robots compute their robot-level scores and aggregate them within each class into the mission-level score of Definition~\ref{def:mission_score}. The mission-level scores collected over the $M_j$ missions form the calibration dataset used to compute the quantile for each class. Since these scores are collected under $\br_j$, we use the score sensitivity bounds to account for the change in the score distribution induced by updating the robustness margins. The quantiles and score sensitivity coefficients are then used to compute $\br_{j+1}$, and the process repeats. Thus, \ac{MAICP} operates across three levels of data: robot-level scores are evaluated over individual trajectories, mission-level scores aggregate the robot-level scores from each mission, and the collection of mission-level scores determines the robustness margin update. The hierarchy of these quantities is illustrated in~\Cref{fig:hero}, while~\Cref{fig:block_diag} shows how \ac{MAICP} fits into the robots' autonomy stack.


Both the mission-level scores and calibration quantiles are order statistics of scalar values held across the robots. We use the same distributed pinball loss optimization algorithm to compute both quantities, with $K_S$ iterations for the mission-level scores and $K_q$ iterations for the calibration quantiles. Since the distributed optimization converges asymptotically, we additionally construct valid upper bounds on the order statistics after a finite number of iterations. The mission-level scores are computed within each subteam, while the calibration quantiles are computed over the connected update graph $\Gcal_j^{\mathrm{up}}$, with each score included once in the corresponding distributed objective. Algorithm~\ref{alg:ma_cp_agent} summarizes one iteration of \ac{MAICP}, using Algorithm~\ref{alg:distributed_order_statistic} for the distributed order statistic computations.


\begin{algorithm}
    \SetKwInOut{Parameters}{Parameters}
    \SetKwInOut{Inputs}{Inputs}
    \caption{\ac{MAICP} at robot $i$}
    \label{alg:ma_cp_agent}
\Parameters{$K_S,K_q,M_j,D,\Rcal_j,\{\alpha_c,\delta_{c,j}^{\mathrm{cal}},\ell_c,w_c\}_{c\in\Ccal},\lambda$}
\Inputs{$j,\br_j,\{\kappa_{c,j}\}_{c\in\Ccal}$, and initial conservative margin $\br_0$}

    \ForEach{$c\in\Ccal$}{
        {Compute $k_{c,j}$ using~\eqref{eq:binomial_calibration_rank}}\;
    }


    \ForEach{$m\in[M_j]$ with $i\in\Vcal^{\mathrm{mis}}(\omega_j^{(m)})$}{
    Execute mission $m$ under $\br_j$\;
    Compute $S_{i,j}^{(m)}$ using~\eqref{eq:nonconformity_score}\;
    \ForEach{$c\in\Ccal$}{
        Compute $\bar S_{c,j}^{(m)}$ within the subteam using Algorithm~\ref{alg:distributed_order_statistic} with rank $K_c$\;
    }
    }
    \ForEach{$c\in\Ccal$}{
        Compute $\bar q_{c,j}^{K_S,K_q}$ over $\Gcal_j^{\mathrm{up}}$ using Algorithm~\ref{alg:distributed_order_statistic} with rank $k_{c,j}$\;
    }
    Construct $\bar\Fcal_j^{K_S,K_q}$ using~\eqref{eq:finite_feasible_set}\;
    \If{$\bar\Fcal_j^{K_S,K_q}\neq\emptyset$}{
        Compute $\br_{j+1}$ using~\eqref{eq:margin_update}\;
    }
    \Else{
        $\br_{j+1}\gets\br_0$\;
    }
    \Return{$\br_{j+1}$}\;
    
\end{algorithm}


\subsection{Iterative Calibration Procedure}

We now describe the three stages of Algorithm~\ref{alg:ma_cp_agent}, which are used to compute the mission-level calibration quantities and update the robustness margins. The distributed computation of the order statistics required by these stages is presented in~\Cref{sec:distributed_order_statistic}.

\subsubsection{Robot-Level Score Computation}

{At iteration $j$ of \ac{MAICP}, the robots execute $M_j$ missions under the fixed robustness margin vector $\br_j$. After each mission, every participating robot computes its realized robot-level score $S_{i,j}^{(m)}$ according to Definition~\ref{def:robot_score}. Although these scores may depend on interactions among the robots during the mission, each robot computes and retains its own score for the distributed mission-level computation.}

\subsubsection{Mission-Level Calibration}
\label{sec:mission_calibration}

After each mission, the participating robots aggregate their robot-level scores within each class as in Definition~\ref{def:mission_score}. Since $K_c=N_c^{\mathrm{mis}}-\ell_c$, at least $K_c$ robots must satisfy their coverage requirement. For mission $m$, let $\bar S_{c,j}^{(m)}$ denote the upper bound computed by Algorithm~\ref{alg:distributed_order_statistic} using rank $K_c$ and $K_S$ distributed iterations, such that $S_{c,j}^{\mathrm{mis},(m)}\leq\bar S_{c,j}^{(m)}$. The calibration dataset for class $c$ consists of the $M_j$ mission-level scores, rather than the individual robot-level scores.

For each class $c\in\Ccal$, let $B_{c,j}\sim\operatorname{Binomial}(M_j,1-\alpha_c)$. We select the calibration quantile rank $k_{c,j}\in[M_j]$ such that
\eqn{
\Pr\left(B_{c,j}\leq k_{c,j}-1\right)\geq1-\delta_{c,j}^{\mathrm{cal}}.
\label{eq:binomial_calibration_rank}
}
Such a rank exists when $\delta_{c,j}^{\mathrm{cal}}\geq(1-\alpha_c)^{M_j}$; we use the smallest admissible rank as in~\eqref{eq:prelim_binomial_rank} with $n=M_j$. Let $q_{c,j}:=S_{c,j,[k_{c,j}]}^{\mathrm{mis}}$ denote the corresponding order statistic of the mission-level scores. Applying Algorithm~\ref{alg:distributed_order_statistic} to $\{\bar S_{c,j}^{(m)}\}_{m=1}^{M_j}$, with each bound included once, produces $\bar q_{c,j}^{K_S,K_q}\geq q_{c,j}$ after $K_q$ distributed iterations.

\subsubsection{Robustness Margin Update}
\label{sec:kappa}

{The quantile $q_{c,j}$ is calibrated using missions executed under $\br_j$, whereas the next missions are executed under $\br_{j+1}$. Since changing the robustness margins can change robot trajectories and distributed algorithm outputs, the score distribution may shift. The score sensitivity bound in Definition~\ref{def:score_sensitivity_bound} accounts for this shift: it is sufficient to require $r_{c,j+1}\geq q_{c,j}+\kappa_{c,j}\Delta_j(\br_{j+1})$ for every class. We therefore use the finite-iteration quantile upper bound $\bar q_{c,j}^{K_S,K_q}$ in the distributed implementation and define the feasible set}
\eqn{
\bar\Fcal_j^{K_S,K_q}:=\left\{\br\in\Rcal_j:\bar q_{c,j}^{K_S,K_q}+\kappa_{c,j}\Delta_j(\br)\leq r_c,\ c\in\Ccal\right\}.
\label{eq:finite_feasible_set}
}
We select the updated robustness margins by solving
\eqn{
\br_{j+1}:=\argmin{\br\in\bar\Fcal_j^{K_S,K_q}}\left(\sum_{c\in\Ccal}w_c r_c+\frac{\lambda}{2}\Delta_j(\br)^2\right),
\label{eq:margin_update}
}
where $w_c>0$ and $\lambda>0$. The first term favors smaller margins, while the second penalizes changes between iterations. The objective is strongly convex, so the minimizer is unique whenever the feasible set is nonempty. Each robot obtains the same quantile upper bounds and score sensitivity coefficients over the connected update graph, and therefore computes the same updated margins. {If the feasible set is empty, Algorithm~\ref{alg:ma_cp_agent} returns $\br_0$.}
\begin{assumption}[Initial Conservative Margin]
\label{ass:initial_margin}
The initial robustness margin vector $\br_0\in\Rcal_j$ satisfies
\eqn{
\mathbb P\left(
S_c^{\mathrm{mis}}(\omega_j^+;\br_0)\leq r_{c,0}
\,\middle|\,\Hcal_j
\right)\geq1-\alpha_c
\label{eq:initial_margin_coverage}
}
almost surely for every class $c\in\Ccal$ and iteration $j\in\N$.
\end{assumption}

\begin{algorithm}[t]
    \SetKwInOut{Parameters}{Parameters}
    \SetKwInOut{Inputs}{Inputs}
    \SetKwInOut{Output}{Output}
    \caption{Distributed Order Statistic}
    \label{alg:distributed_order_statistic}
    \Parameters{$W,\{\gamma_\ell\}_{\ell=0}^{K-1}$}
\Inputs{$\Gcal',\{(s,a_s):s\in\Jcal_i\},L,k,K$}
    \Output{$\bar a_{[k]}^K$}

    $\tau_k\gets(k-\frac{1}{2})/L$\;
    Initialize $q_i^0\in\R$\;

    \For{$\ell=0,\ldots,K-1$}{
        Exchange $q_i^\ell$ with $h\in{\Ncal_i'}$\;
        Compute $g_i^\ell$ using~\eqref{eq:quantile_subgradient}\;
        Update $q_i^{\ell+1}$ using~\eqref{eq:distributed_quantile_update}\;
    }

    $a_i^{\max}\gets\max\left(\{-\infty\}\cup\{a_s:s\in\Jcal_i\}\right)$\;
    Flood $(a_i^{\max},q_i^K)$ to obtain $a_{\max}$ and $q_{\max}^K$\;
    $y\gets\min\{a_{\max},q_{\max}^K\}$\;

    \Repeat{$N^{\leq}(y)\geq k$}{
        Compute $N_i^{\leq}(y)$ and $R_i(y)$\;
{        Flood $(i,N_i^{\leq}(y))$ and $R_i(y)$ to obtain $N^{\leq}(y)$ and $R(y)$\;}
        \If{$N^{\leq}(y)<k$}{
            $y\gets R(y)$\;
        }
    }

    $\bar a_{[k]}^K\gets y$\;
    \Return{$\bar a_{[k]}^K$}\;
\end{algorithm}

\subsection{Distributed Order Statistic Computation}
\label{sec:distributed_order_statistic}

The mission-level calibration stage requires two distributed order statistic computations: one to aggregate robot-level scores within each mission and another to compute a calibration quantile across missions. We use the same algorithm for both computations. Let $\{a_1,\ldots,a_L\}$ denote the scalar values and $k\in[L]$ the desired rank. For mission-level scores, $L=N_c^{\mathrm{mis}}$, $k=K_c$, and $K=K_S$, and for calibration quantiles, $L=M_j$, $k=k_{c,j}$, and $K=K_q$.

Let $\Gcal'=(\Vcal',\Ecal')$ denote the connected graph used for either computation, with neighbor set $\Ncal_i':=\{h\in\Vcal':(i,h)\in\Ecal'\}$. For mission-level scores, $\Vcal'=\Vcal^{\mathrm{mis}}(\omega_j^{(m)})$, while calibration quantiles use $\Gcal'=\Gcal_j^{\mathrm{up}}$. Each robot $i\in\Vcal'$ holds the values indexed by $\Jcal_i\subseteq[L]$, and the sets $\{\Jcal_i\}_{i\in\Vcal'}$ partition $[L]$, ensuring that each value is counted once.

Using the pinball loss $\ell_\tau(v):=\tau[v]_++(1-\tau)[-v]_+$ for $\tau\in(0,1)$, set $\tau_k:=(k-\frac12)/L$ and $F_k(q):=\sum_{s=1}^L\ell_{\tau_k}(a_s-q)$. The unique minimizer is $a_{[k]}$, including when values are repeated: since $\tau_k L=k-\frac12$, the subgradients of $F_k$ are negative below $a_{[k]}$ and positive above it. The objective is separable as $F_k(q)=\sum_{i\in\Vcal'}\phi_i(q)$, where $\phi_i(q):=\sum_{s\in\Jcal_i}\ell_{\tau_k}(a_s-q)$. Robot $i$ computes the local subgradient
\eqn{g_i^\ell:=\sum_{s\in\Jcal_i}\left(\mathbf{1}_{\{a_s<q_i^\ell\}}-\tau_k\right)\in\partial\phi_i(q_i^\ell).\label{eq:quantile_subgradient}}

Here, $\partial\phi_i(q)$ denotes the convex subdifferential of $\phi_i$ at $q$. When $\Jcal_i=\emptyset$, we set $\phi_i\equiv0$ and $g_i^\ell=0$. With symmetric, doubly stochastic weights $W=[w_{ih}]_{i,h\in\Vcal'}$ satisfying $w_{ii}>0$, $w_{ih}>0$ on edges, and $w_{ih}=0$ for non-neighboring distinct robots, the distributed subgradient iteration is
\eqn{
q_i^{\ell+1}=\sum_{h\in\Ncal_i'\cup\{i\}}w_{ih}q_h^\ell-\gamma_\ell g_i^\ell,
\label{eq:distributed_quantile_update}
}
where $\gamma_\ell>0$, $\sum_{\ell=0}^\infty\gamma_\ell=\infty$, and $\sum_{\ell=0}^\infty\gamma_\ell^2<\infty$.

A finite number $K$ of distributed iterations does not guarantee that $q_i^K$ upper bounds $a_{[k]}$. We therefore compute $a_{\max}:=\max_{s\in[L]}a_s$ and $q_{\max}^K:=\max_{i\in\Vcal'}q_i^K$ using flooding~\cite{lynch1996distributed}, in which robots forward received information to their neighbors, and initialize a candidate $y:=\min\{a_{\max},q_{\max}^K\}$. For this candidate, robot $i$ computes the local rank count $N_i^{\leq}(y):=\sum_{s\in\Jcal_i}\mathbf{1}_{\{a_s\leq y\}}$ and next larger value $R_i(y):=\min(\{+\infty\}\cup\{a_s:s\in\Jcal_i,\ a_s>y\})$. Flooding determines $N^{\leq}(y):=\sum_iN_i^{\leq}(y)$ and $R(y):=\min_iR_i(y)$.

{The condition $N^{\leq}(y)\geq k$ holds when $y\geq a_{[k]}$. Otherwise, $R(y)$ is the smallest observed value above $y$ and satisfies $R(y)\leq a_{[k]}$. Updating the candidate to $R(y)$ strictly increases $N^{\leq}(y)$, so Algorithm~\ref{alg:distributed_order_statistic} terminates after at most $k$ candidate updates with $a_{[k]}\leq\bar a_{[k]}^K$ for every finite $K$. Moreover, as the distributed subgradient iterates converge to $a_{[k]}$, the recovered bound satisfies $\bar a_{[k]}^K\to a_{[k]}$ as $K\to\infty$. To avoid counting an input more than once along different paths, robots include their identifiers when flooding rank counts. Each flooding operation requires at most $\operatorname{diam}(\Gcal')$ communication rounds, where $\operatorname{diam}(\Gcal')$ is the maximum shortest-path distance between robots, with up to $\abs{\Vcal'}$ counts and identifiers per message.}

\section{Theoretical Analysis}
\label{sec:theory}
Next, we provide theoretical analysis for the coverage guarantee of \ac{MAICP} using the iteratively-updated robustness margins. We first establish the distributed order statistic bounds and the calibration guarantee under the current margins. Then, we bound changes in the robot-level scores and combine these results to establish coverage under the updated margins.

\subsection{Distributed Quantile Computation}
Under the connected communication graph and step size conditions in~\Cref{sec:distributed_order_statistic}, {the distributed subgradient descent iterates satisfy}
\eqnN{
    \lim_{\ell\to\infty}
    q_i^\ell
    =
    a_{[k]},
    \qquad
    i\in{\Vcal'},
}
since $a_{[k]}$ is the unique minimizer of $F_k$~\cite{nedic2010constrained}.

\begin{prop}[Order Statistic Upper Bound]
\label{prop:finite_order_statistic}
{For finite input scores $a_1,\ldots,a_L\in\R$ and every finite $K\in\{0\}\cup\N$, Algorithm~\ref{alg:distributed_order_statistic} satisfies}
\eqn{
    a_{[k]}
    \leq
    \bar a_{[k]}^K
    \leq
    a_{\max}.
    \label{eq:finite_order_statistic_bound}
}
{Moreover, $\bar a_{[k]}^K\to a_{[k]}$ as $K\to\infty$. Let $D_j$ bound the diameters of the connected communication graphs used at iteration $j$. Each distributed computation then requires at most $K+D_j(k+2)$ communication rounds.}
\end{prop}

\begin{proof}
By construction, $
    N^{\leq}(y)
    =
    \sum_{s=1}^{L}
    \mathbf{1}_{\{a_s\leq y\}},$
and therefore $
    N^{\leq}(y)\geq k
    \quad\Longleftrightarrow\quad
    y\geq a_{[k]}.$
If $N^{\leq}(y)<k$, then $y<a_{[k]}$ and, since $a_{[k]}$ is contained in $\{a_s:a_s>y\}$, $
    y
    <
    R(y)
    \leq
    a_{[k]}.$
Thus, each update $y\gets R(y)$ strictly increases $N^{\leq}(y)$ and the procedure terminates after at most $k$ updates. If the initial value $y_0^K:=\min\{a_{\max},q_{\max}^K\}$ satisfies $y_0^K<a_{[k]}$, the procedure terminates at $a_{[k]}$; otherwise, it returns $y_0^K$. Hence $
    \bar a_{[k]}^K
    =
    \max
    \left\{
        a_{[k]},
        y_0^K
    \right\},$
{and therefore~\eqref{eq:finite_order_statistic_bound} holds.} Since $q_{\max}^K\to a_{[k]}$, we have $y_0^K\to a_{[k]}$ and therefore $\bar a_{[k]}^K\to a_{[k]}$. The $K$ distributed subgradient iterations require $K$ communication rounds. Computing $(a_{\max},q_{\max}^K)$ requires one flooding operation. Evaluating the initial rank condition and the rank conditions following at most $k$ {candidate updates} requires at most $k+1$ additional flooding operations. Since each flooding operation requires at most $D_j$ rounds, the total number of communication rounds is at most $K+D_j(k+2)$.

\end{proof}


\subsection{Calibration Under the Current Robustness Margins}
We next apply the calibration-conditional coverage guarantee to the mission-level scores collected under $\br_j$. Under Assumption~\ref{ass:mission_sampling}, each mission provides one independent calibration data point, while the robot-level scores within a mission may be dependent.
\begin{lemma}[Mission-Level Calibration]
\label{lem:mission_calibration}
{Suppose Assumption~\ref{ass:mission_sampling} holds and the rank $k_{c,j}$ satisfies~\eqref{eq:binomial_calibration_rank}. Then}
\eqn{
\begin{aligned}
\mathbb P
\Bigg[
    \mathbb P
    \left(
        S_c^{\mathrm{mis}}
        \left(
            \omega_j^+;
            \br_j
        \right)
        \leq
        q_{c,j}
        \,\middle|\,
        \Ical_j
    \right)
    \geq
    1-\alpha_c
    \,\Bigg|\,
    \Hcal_j
\Bigg]
\geq
1-\delta_{c,j}^{\mathrm{cal}}.
\end{aligned}
\label{eq:current_margin_coverage}
}
\end{lemma}

\begin{proof}
Conditional on $\Hcal_j$, Assumption~\ref{ass:mission_sampling} implies that the $M_j$ mission scores and the next mission score are i.i.d. and that the mission is conditionally independent of $\Dcal_j$. {The bound~\eqref{eq:current_margin_coverage} follows from~\cite[Prop. 2b]{vovk2012conditional}, using the rank specified by~\eqref{eq:binomial_calibration_rank}.}
\end{proof}

\subsection{Score Sensitivities}

Our main coverage result requires {a class-dependent score sensitivity coefficient $\kappa_{c,j}$} that bounds the change in the mission-level score as the robustness margins are updated, as stated in Definition~\ref{def:score_sensitivity_bound}. This coefficient may be obtained directly from an application-specific analysis. We next provide one sufficient construction by first bounding the sensitivity of each robot-level score to the robustness margins and then showing that a class-wide bound on the robot-level sensitivities also bounds the mission-level score.

\subsubsection{Robot-Level Score Sensitivity for Distributed Algorithms}

We bound the change in the robot-level scores by comparing two closed-loop executions under $\br$ and $\br_j$ with the same mission realization $\omega$. {Let $0=t_0<\cdots<t_{N_t}=T$ denote comparison times used to bound deviations relevant to the robot-level scores.} At time $t_\ell$, let $d_i^\ell\in\Rnonneg$ measure the difference in selected physical or internal algorithm states of robot $i$ between the two executions, with $d_i^0=0$ since their initial states coincide. We define $d^\ell:=\operatorname{col}(d_1^\ell,\ldots,d_N^\ell)$ and $\bar d:=\operatorname{col}(\bar d_1,\ldots,\bar d_N)$, where $\bar d_i:=\max_{\ell=0,\ldots,N_t}d_i^\ell$. {When the score evaluation times differ between the two executions, their effect is included in the robot-level score bound below. We assume the following bounds on the propagation of these deviations and their effect on the robot-level scores.}

\begin{assumption}[Distributed Algorithm Sensitivity Bounds]
\label{ass:sensitivity}
At each iteration $j$, the propagation of these deviations and their effect on the robot-level scores are bounded using $\Ical_j$-measurable coefficients $A_{\ell,j}\in\Rnonneg^{N\times N}$, $b_{\ell,j},L_{i,j}\in\Rnonneg^N$, and $L_{i,j}^{r}\in\Rnonneg$. Conditional on $\Ical_j$, for almost every $\omega$ these bounds hold for all $\br\in\Rcal_j$:
\eqn{
d^{\ell+1}\preceq A_{\ell,j}d^\ell+b_{\ell,j}\Delta_j(\br),\quad \ell=0,\ldots,N_t-1,
\label{eq:incremental_sensitivity}
}
and, for every $i\in\Vcal$,
\eqn{
S_i(\omega;\br)-S_i(\omega;\br_j)\leq L_{i,j}^\top\bar d+L_{i,j}^{r}\Delta_j(\br).
\label{eq:score_sensitivity}
}
\end{assumption}

\begin{prop}[Robot-Level Score Sensitivity]
\label{prop:robot_kappa}
Suppose Assumptions~\ref{ass:well_posed} and~\ref{ass:sensitivity} hold. Define $G_{0,j}=0$ and $G_{\ell+1,j}=A_{\ell,j}G_{\ell,j}+b_{\ell,j}$ for $\ell=0,\ldots,N_t-1$. Let $\bar G_j\in\Rnonneg^N$ have components $[\bar G_j]_i:=\max_{\ell=0,\ldots,N_t}[G_{\ell,j}]_i$. Then conditional on $\Ical_j$ and for almost every mission realization $\omega$, the robot-level score sensitivity coefficient $\kappa_{i,j}:=L_{i,j}^{r}+L_{i,j}^\top\bar G_j$ satisfies 
\eqn{
S_i(\omega;\br)\leq S_i(\omega;\br_j)+\kappa_{i,j}\Delta_j(\br)
\label{eq:robot_score_shift}
}
simultaneously for every $\br\in\Rcal_j$ and $i\in\Vcal$.
\end{prop}

\begin{proof}
Since the paired executions have identical initial conditions, $d^0=0$. Suppose $d^\ell\preceq G_{\ell,j}\Delta_j(\br)$. Using~\eqref{eq:incremental_sensitivity} and the nonnegativity of $A_{\ell,j}$, we have
\eqnN{
\begin{aligned}
d^{\ell+1}&\preceq A_{\ell,j}d^\ell+b_{\ell,j}\Delta_j(\br)\\
&\preceq (A_{\ell,j}G_{\ell,j}+b_{\ell,j})\Delta_j(\br)=G_{\ell+1,j}\Delta_j(\br).
\end{aligned}
}
Hence, by induction, $d^\ell\preceq G_{\ell,j}\Delta_j(\br)$ for all $\ell\in\{0,\ldots,N_t\}${; consequently, $\bar d\preceq\bar G_j\Delta_j(\br)$.} {Substituting this bound into~\eqref{eq:score_sensitivity} and using the nonnegativity of $L_{i,j}$, the score difference satisfies}
\eqnN{
\begin{aligned}
S_i(\omega;\br)-S_i(\omega;\br_j)
\leq (L_{i,j}^\top\bar G_j+L_{i,j}^{r})\Delta_j(\br)=\kappa_{i,j}\Delta_j(\br).
\end{aligned}
}
\end{proof}

\subsubsection{Mission-Level Sensitivity}

We next show that a class sensitivity coefficient that bounds the change in every participating robot-level score also bounds the change in the corresponding mission-level score.

\begin{prop}[Mission-Level Score Sensitivity]
\label{prop:class_kappa}
Fix class $c\in\Ccal$ and iteration $j$. Suppose $\kappa_{c,j}\in\Rnonneg$ is $\Ical_j$-measurable and, conditional on $\Ical_j$, for almost every mission realization $\omega$,
\eqn{
    S_i
    \left(
        \omega;
        \br
    \right)
    \leq
    S_i
    \left(
        \omega;
        \br_j
    \right)
    +
    \kappa_{c,j}
    \Delta_j
    \left(
        \br
    \right)
    \label{eq:class_robot_score_shift}
}
holds for every $i\in\Vcal_c^{\mathrm{mis}}(\omega)$ and simultaneously for every $\br\in\Rcal_j$. Then
\eqn{
    S_c^{\mathrm{mis}}
    \left(
        \omega;
        \br
    \right)
    \leq
    S_c^{\mathrm{mis}}
    \left(
        \omega;
        \br_j
    \right)
    +
    \kappa_{c,j}
    \Delta_j
    \left(
        \br
    \right),
    \label{eq:mission_score_shift_from_robot}
}
and therefore $\kappa_{c,j}$ satisfies Definition~\ref{def:score_sensitivity_bound}.
\end{prop}

\begin{proof}
{For $\omega$ and $\br\in\Rcal_j$, {the mission-level inequality~\eqref{eq:mission_score_shift_from_robot} follows by taking the $K_c$th order statistic in~\eqref{eq:class_robot_score_shift} and using its monotonicity.}}
\end{proof}

In particular, Proposition~\ref{prop:robot_kappa} satisfies the condition of Proposition~\ref{prop:class_kappa} for any $\kappa_{c,j}$ satisfying $\kappa_{c,j}\geq\kappa_{i,j}$ for every $i\in\Vcal_c$.

\subsection{Coverage After Margin Update}

We now combine the mission-level calibration result, the score sensitivity bound, and the finite distributed order statistic computations to establish the coverage guarantee for \ac{MAICP}. In particular, for any $\Ical_j$-measurable quantile bound $\hat q_{c,j}\geq q_{c,j}$,
\eqn{
\hat q_{c,j}+\kappa_{c,j}\Delta_j(\br_{j+1})\leq r_{c,j+1},\qquad c\in\Ccal,
\label{eq:general_margin_constraint}
}
is sufficient to transfer the calibration guarantee to the updated margin. \Cref{alg:ma_cp_agent}~enforces this condition through~\eqref{eq:finite_feasible_set} when the feasible set is nonempty and otherwise returns $\br_0$.

\begin{theorem}[Conditional Coverage of Distributed \ac{MAICP}]
\label{th:maicp_coverage}
Suppose Assumptions~\ref{ass:mission_sampling}, \ref{ass:score_sensitivity}, and~\ref{ass:initial_margin} hold, the rank $k_{c,j}$ satisfies~\eqref{eq:binomial_calibration_rank} for every $c\in\Ccal$, and the conditions of Proposition~\ref{prop:finite_order_statistic} hold for both distributed order statistic computations. Then, for every finite $K_S,K_q\in\{0\}\cup\N$, \Cref{alg:ma_cp_agent}~returns $\br_{j+1}\in\Rcal_j$. For each class $c\in\Ccal$, let $\Ecal_{c,j}^{\mathrm{mis}}:=\{S_c^{\mathrm{mis}}(\omega_j^+;\br_{j+1})\leq r_{c,j+1}\}$. Then,
\eqn{
\mathbb P\left[\mathbb P(\Ecal_{c,j}^{\mathrm{mis}}\mid\Ical_j)\geq1-\alpha_c\,\middle|\,\Hcal_j\right]\geq1-\delta_{c,j}^{\mathrm{cal}}.
\label{eq:maicp_conditional_coverage}
}
Furthermore, if $\sum_{c\in\Ccal}\delta_{c,j}^{\mathrm{cal}}<1$, all classwise conditional coverage guarantees hold simultaneously with
\eqn{
\begin{aligned}
\mathbb P\Bigg[
\bigcap_{c\in\Ccal}
\left\{\mathbb P(\Ecal_{c,j}^{\mathrm{mis}}\mid\Ical_j)\geq1-\alpha_c\right\}
\,\Bigg|\,\Hcal_j\Bigg]
\geq1-\sum_{c\in\Ccal}\delta_{c,j}^{\mathrm{cal}}.
\end{aligned}
\label{eq:simultaneous_class_coverage}
}
\end{theorem}

\begin{proof}
For each $c\in\Ccal$, define the calibration event
\eqnN{
\Ecal_{c,j}^{\mathrm{cal}}
:=
\left\{
\mathbb P\left(
S_c^{\mathrm{mis}}(\omega_j^+;\br_j)\leq q_{c,j}
\,\middle|\,\Ical_j
\right)\geq1-\alpha_c
\right\}.
}
{By Lemma~\ref{lem:mission_calibration}, $\mathbb P(\Ecal_{c,j}^{\mathrm{cal}}\mid\Hcal_j)\geq1-\delta_{c,j}^{\mathrm{cal}}$.} If $\bar\Fcal_j^{K_S,K_q}\neq\emptyset$, Algorithm~\ref{alg:ma_cp_agent} returns $\br_{j+1}$ satisfying~\eqref{eq:general_margin_constraint} with $\hat q_{c,j}=\bar q_{c,j}^{K_S,K_q}\geq q_{c,j}$, where the inequality follows from Proposition~\ref{prop:finite_order_statistic} applied to both computations and the monotonicity of order statistics. By Definition~\ref{def:score_sensitivity_bound}, this implies
\eqnN{
\left\{
S_c^{\mathrm{mis}}(\omega_j^+;\br_j)\leq q_{c,j}
\right\}
\subseteq
\Ecal_{c,j}^{\mathrm{mis}}
}
for almost every $\omega_j^+$. Hence, on $\Ecal_{c,j}^{\mathrm{cal}}$, $\mathbb P(\Ecal_{c,j}^{\mathrm{mis}}\mid\Ical_j)\geq1-\alpha_c$.

If $\bar\Fcal_j^{K_S,K_q}=\emptyset$, Algorithm~\ref{alg:ma_cp_agent} returns $\br_{j+1}=\br_0$. By Assumption~\ref{ass:mission_sampling}, the post-calibration mission is conditionally independent of $\Dcal_j$ given $\Hcal_j$. {Since $\br_0$ is $\Hcal_j$-measurable, the bound in Assumption~\ref{ass:initial_margin} applies:}
\eqnN{
\mathbb P(\Ecal_{c,j}^{\mathrm{mis}}\mid\Ical_j)
=
\mathbb P\left(
S_c^{\mathrm{mis}}(\omega_j^+;\br_0)\leq r_{c,0}
\,\middle|\,\Hcal_j
\right)\geq1-\alpha_c.
}

Thus, in either case, $\mathbb P(\Ecal_{c,j}^{\mathrm{mis}}\mid\Ical_j)\geq1-\alpha_c$ on $\Ecal_{c,j}^{\mathrm{cal}}$, which proves~\eqref{eq:maicp_conditional_coverage}. {The simultaneous class coverage bound~\eqref{eq:simultaneous_class_coverage} follows by applying the union bound to $\{\Ecal_{c,j}^{\mathrm{cal}}\}_{c\in\Ccal}$ conditional on $\Hcal_j$.}
\end{proof}

Theorem~\ref{th:maicp_coverage} establishes~\eqref{eq:desired_coverage} for the distributed implementation at every finite $K_S,K_q\in\{0\}\cup\N$. In particular, assigning $\alpha_c=\alpha_{\mathrm{joint}}/|\Ccal|$ and $\delta_{c,j}^{\mathrm{cal}}=\delta_{\mathrm{joint}}/|\Ccal|$ provides joint mission coverage of at least $1-\alpha_{\mathrm{joint}}$ with calibration confidence at least $1-\delta_{\mathrm{joint}}$.

\begin{remark}[Coverage Guarantee for a Uniformly Selected Robot]
\label{rem:uniform}
Suppose the mission-level guarantee holds for class $c$, and let $I_c$ denote a robot selected uniformly from $\Vcal_c^{\mathrm{mis}}(\omega_j^+)$. Since $S_c^{\mathrm{mis}}\leq r_c$ implies that at least $K_c=N_c^{\mathrm{mis}}-\ell_c$ participating robots satisfy their robot-level coverage requirement, it follows that
\eqn{
    \mathbb P
    \left(
        S_{I_c}
        \left(
            \omega_j^+;
            \br_{j+1}
        \right)
        \leq
        r_{c,j+1}
        \,\middle|\,
        \Ical_j
    \right)
    \geq
    (1-\alpha_c)
    \frac{K_c}{N_c^{\mathrm{mis}}}.
}
Thus, the guarantee for a uniformly selected participating robot is at least $(1-\alpha_c)(1-\ell_c/N_c^{\mathrm{mis}})$. {By Theorem~\ref{th:maicp_coverage}, this bound holds with conditional probability at least $1-\delta_{c,j}^{\mathrm{cal}}$.}
\end{remark}

\section{Simulation Case Studies}
\label{sec:simulation}

We consider cooperative payload transport, distributed formation control with collision avoidance, and distributed target tracking as separate applications of \ac{MAICP}. The applications use different robot-level scores and introduce the class-dependent robustness margins into different controllers and estimators, while sharing the mission-level score and robustness margin update in~\Cref{sec:method}. Finally, we examine the distributed order statistic computation under different communication graph topologies. See our public code repository for implementation details\footnote{\scriptsize\url{https://github.com/dcherenson/maicp}}.

\begin{figure}[t]
    \centering
    \includegraphics[width=\linewidth]{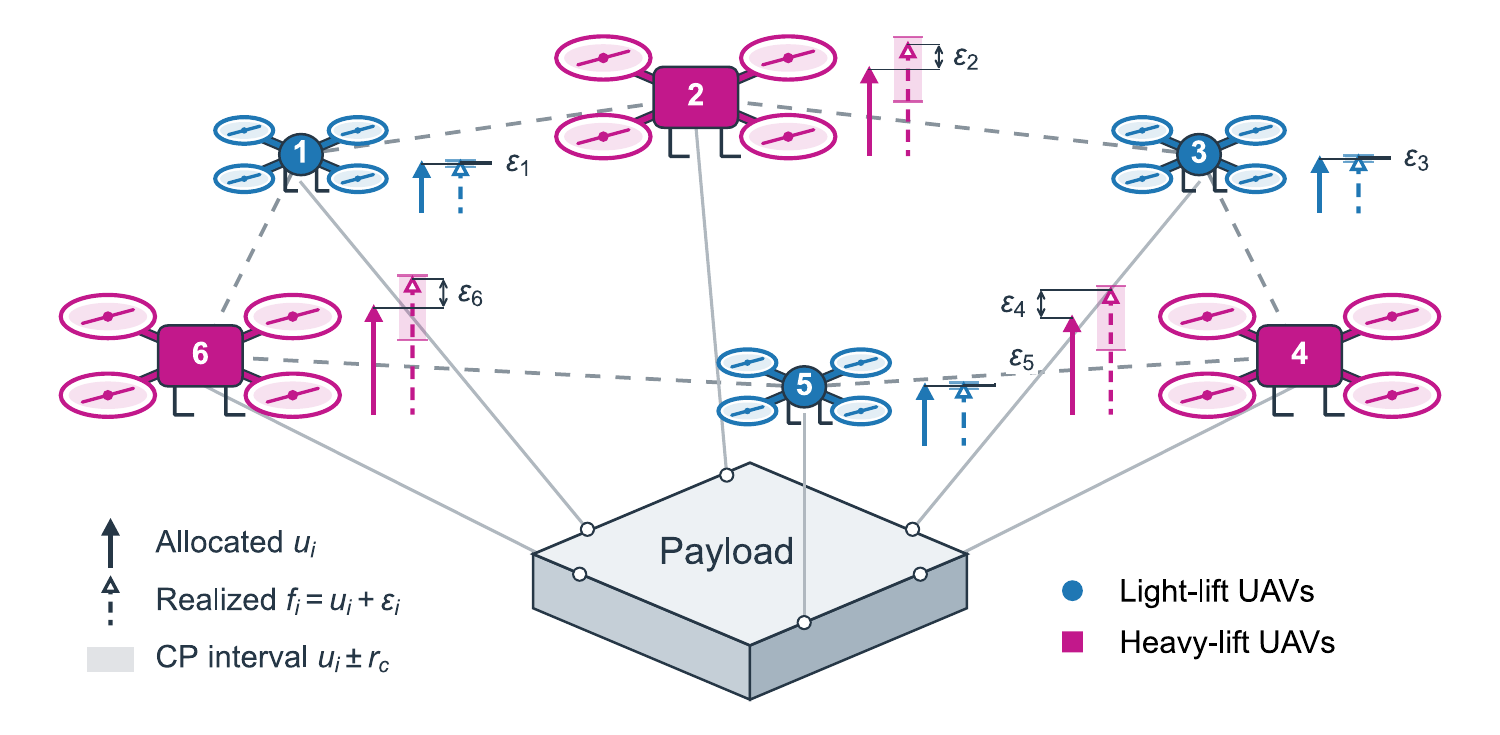}
\caption{Cooperative payload force allocation with three light-lift and three heavy-lift UAVs (dashed lines indicate communication links). The realized force is $f_i=u_i+\epsilon_i$, where $\epsilon_i$ is the unknown actuator force residual, and $u_i\pm r_{C(i)}$ defines the class-dependent robustness interval.}
    \label{fig:payload_cartoon}
\end{figure}

\subsection{Cooperative Payload Force Allocation}
\label{sec:sim_payload}

{We consider cooperative payload load sharing using six unmanned aerial vehicles (UAVs) belonging to two classes, with three robots per class.} Each robot has a regulated vertical position error $e_i$, velocity error $v_i$, and applied load force $u_i$. We model the local compliance dynamics by
\eqn{
\dot e_i=v_i, \quad
m_i\dot v_i=-k_i^p e_i-k_i^v v_i-(u_i-u_i^0+\epsilon_i),
\label{eq:sim_payload_dynamics}
}
{where $m_i>0$ is the effective mass, $k_i^p,k_i^v>0$ are the position and velocity compliance gains, $u_i^0$ denotes a prescribed nominal load force, and $\epsilon_i$ is the unknown force residual.} 

At sampling time $t_k=kh$, the robots distribute the prescribed force $F_k=21+1.2\sin(2\pi(k+1/2)/100)\,\mathrm N$, with sampling period $h=0.1\,\mathrm s$ and mission horizon $T=10\,\mathrm s$. They seek to minimize $\sum_i u_i^2/(2\varpi_i)$ subject to $\sum_i u_i=F_k$, where the positive weights are $\varpi_i(x_i,\br):=a_i\exp(-\mu_i r_{C(i)}+\sigma_i\tanh(e_i/s_i))$, $x_i:=\operatorname{col}(e_i,v_i)$, $\sigma_i=0.08$, and $s_i=1\,\mathrm m$. We use $(a_i)=(1.08,2.34,1.20,2.60,1.32,2.86)$ and $\mu_i=0.10$ and $0.18\,\mathrm N^{-1}$ for classes 1 and 2, respectively. {The robots communicate over a fixed connected ring with adjacency matrix $A_{\Gcal}$ and graph Laplacian $L_{\Gcal}:=\operatorname{diag}(A_{\Gcal}\mathbf1_6)-A_{\Gcal}$, where $\mathbf1_6$ is the vector of ones. They execute $K_u=5$ distributed optimization iterations,}
\eqn{
u_k^{\ell+1}&=\left(I_6-\gamma L_{\Gcal}D_\varpi(x_k,\br)^{-1}\right)u_k^\ell,\quad
D_\varpi:=\operatorname{diag}(\varpi_i),
\label{eq:sim_payload_allocation}
}
{with $\gamma=0.1$.} Each robot starts from {$u_{i,k}^0=F_k a_i/\sum_{s=1}^{6}a_s$} and applies $u_{i,k}=u_{i,k}^{K_u}$. Hence the optimization is terminated after five communication rounds, while the total commanded force remains $F_k$. The nominal force in~\eqref{eq:sim_payload_dynamics} is $u_i^0=u_{i,k}^0$ during sample $k$. See~\Cref{fig:payload_cartoon} for an illustration of the payload mission setup.

We consider the force residual
\eqn{
{\epsilon_{i,k}=\theta_i u_{i,k}+\phi_i\sum_{s=1}^{6}\widetilde A_{is}u_{s,k}+\zeta_i\tanh(\frac{e_{i,k}}{s_i})+\eta_{i,k},}
\label{eq:sim_payload_error}
}
{where $\widetilde A:=\operatorname{diag}(A_{\Gcal}\mathbf1_6)^{-1}A_{\Gcal}$ is the row-normalized adjacency matrix,} $\theta_i$, $\phi_i$, and $\zeta_i$ are constant during each mission, and $\eta_{i,k}$ is an additive disturbance. We assume $\abs{\theta_i}\leq\bar\theta_{C(i)}$, $\abs{\phi_i}\leq\bar\phi_{C(i)}$, and $\abs{\zeta_i}\leq\bar\zeta$, with $(\bar\theta_1,\bar\theta_2)=(0.08,0.25)$, $(\bar\phi_1,\bar\phi_2)=(0.03,0.08)$, and $\bar\zeta=0.08\,\mathrm N$. The additive disturbance satisfies $\abs{\eta_{i,k}}\leq0.034\,\mathrm N$, and initial position and velocity errors are bounded by $0.05\,\mathrm m$ and $0.02\,\mathrm{m/s}$. Paired executions under different margin vectors use the same exogenous realization, including the unknown coefficients, additive disturbances, and initial states. The local position and velocity are assumed known, the graph is fixed, and communication is synchronous and reliable. {The force bounds used in the sensitivity comparison are constructed from the allocation recursion below.} The force residuals are held constant within each sampling interval. Taking $z_i=u_i$, $z_i^\star=u_i+\epsilon_i$, and $\rho_c(z_i,z_i^\star)=\abs{z_i-z_i^\star}$ in Definition~\ref{def:robot_score}, the robot-level score is $S_i(\omega;\br)=\max_{k=0,\ldots,99}\abs{\epsilon_{i,k}(\omega;\br)}$.

{We next derive the score sensitivity coefficients by bounding changes in the five allocation iterations and propagating them through the payload dynamics. For the specified parameters, these bounds are $\kappa_{1,j}\approx0.106$ and $\kappa_{2,j}\approx0.324$.} {Let $E\in\R^{6\times2}$ satisfy $E_{ic}=\mathbf1_{\{C(i)=c\}}$.} Since $\br\in\Rcal_j=[0,2]^2\,\mathrm N$ and $\abs{\tanh(e_i/s_i)}\leq1$, the allocation weights satisfy $\underline\varpi_i\leq\varpi_i\leq\overline\varpi_i$, where $\underline\varpi_i:=a_i\exp(-2\mu_i-\sigma_i)$ and $\overline\varpi_i:=a_i\exp(\sigma_i)$. The comparison matrix for a communication round is
\eqn{
[A^u]_{ih}=\begin{cases}
1-\gamma\deg(i)/\overline\varpi_i,&h=i,\\
\gamma [A_{\Gcal}]_{ih}/\underline\varpi_h,&h\neq i.
\end{cases}
\label{eq:sim_payload_Au}
}
The chosen step size ensures $\gamma\deg(i)/\underline\varpi_i\leq1$, so the actual distributed iteration preserves nonnegative forces and their sum. Let $U_i^\ell$ denote an upper bound on the force at communication round $\ell$, initialized with {$U_i^0=F_{\max}a_i/\sum_{s=1}^{6}a_s$} and updated by $U^{\ell+1}=\min\{F_{\max}\mathbf1_6,A^uU^\ell\}$ componentwise, where $F_{\max}=22.2\,\mathrm N$. {Let $\epsilon_k:=\operatorname{col}(\epsilon_{1,k},\ldots,\epsilon_{6,k})$. We use $\Delta e_{i,k}$ and $\Delta\epsilon_k$ to denote the differences in position error and the force residual vector, respectively, between executions under $\br$ and $\br_j$.}

For the state sensitivity, let $F_i=\bmat{0&1\\-k_i^p/m_i&-k_i^v/m_i}$, $P_i=\exp(hF_i)$, and $Q_i=\int_0^h\exp(F_i\tau)(0,-1/m_i)^\top d\tau$. Since $F_i$ is Hurwitz, choose $\alpha_i>0$ such that $F_i+\alpha_i I_2$ is Hurwitz, and let $H_i\succ0$ solve $(F_i+\alpha_iI_2)^\top H_i+H_i(F_i+\alpha_iI_2)=-I_2$, with $T_i^\top T_i=H_i$. We define $d_{i,k}:=\norm{T_i(x_{i,k}(\omega;\br)-x_{i,k}(\omega;\br_j))}_2$ and $d_k:=\operatorname{col}(d_{1,k},\ldots,d_{6,k})$. {For this fixed transformation, define $\rho_i:=e^{-\alpha_i h}$, $\beta_i:=\norm{T_iQ_i}_2$, and $c_i^p:=\norm{(1,0)T_i^{-1}}_2$, which satisfy} $\norm{T_iP_iT_i^{-1}}_2\leq\rho_i$ and $\abs{\Delta e_{i,k}}\leq c_i^p d_{i,k}$.

Let $d_{u,k}^\ell:=\abs{u_k^\ell(\omega;\br)-u_k^\ell(\omega;\br_j)}$, with vector and matrix absolute values taken entrywise. Define $\Lambda^\ell:=\gamma\abs{L_{\Gcal}}\operatorname{diag}(U_i^\ell/\underline\varpi_i)$, $V^\ell:=\Lambda^\ell\operatorname{diag}(\sigma_i c_i^p/s_i)$, and $[b^\ell]_i:=\norm{[\Lambda^\ell\operatorname{diag}(\mu_i)E]_{i,:}}_2$. {The deviations between the distributed allocation iterations under $\br$ and $\br_j$ satisfy}
\eqn{
d_{u,k}^{\ell+1}\preceq A^u d_{u,k}^\ell+V^\ell d_k+b^\ell\Delta_j(\br).
\label{eq:sim_payload_internal_comparison}
}


Here $\Delta_j(\br)=\norm{\br-\br_j}_2$. Since $d_{u,k}^0=0$, let $C_0=g_0=0$ and define $C_{\ell+1}=A^uC_\ell+V^\ell$ and $g_{\ell+1}=A^ug_\ell+b^\ell$. After $K_u=5$ communication rounds, $d_{u,k}^{K_u}\preceq Cd_k+g\Delta_j(\br)$, where $C:=C_{K_u}$ and $g:=g_{K_u}$. Define $H_u:=\operatorname{diag}(\bar\theta_{C(i)})+\operatorname{diag}(\bar\phi_{C(i)})\widetilde A$ and $H_x:=\operatorname{diag}(\bar\zeta c_i^p/s_i)$. {For the same $\omega$, the residual differences satisfy $\abs{\Delta\epsilon_k}\preceq H_u d_{u,k}^{K_u}+H_xd_k$.} The additive disturbance cancels, while the state-dependent residual contributes through $H_x$. With $R:=\operatorname{diag}(\rho_i)$ and $B:=\operatorname{diag}(\beta_i)$, we arrive at the recursion in~\eqref{eq:incremental_sensitivity}, with
\eqn{
\begin{aligned}
d_{k+1}&\preceq A_{k,j}d_k+b_{k,j}\Delta_j(\br),\\
A_{k,j}&:=R+B\left((I_6+H_u)C+H_x\right),\\
b_{k,j}&:=B(I_6+H_u)g.
\end{aligned}
\label{eq:sim_payload_physical_comparison}
}

\begin{figure}[t]
    \centering
    \includegraphics[width=\linewidth]{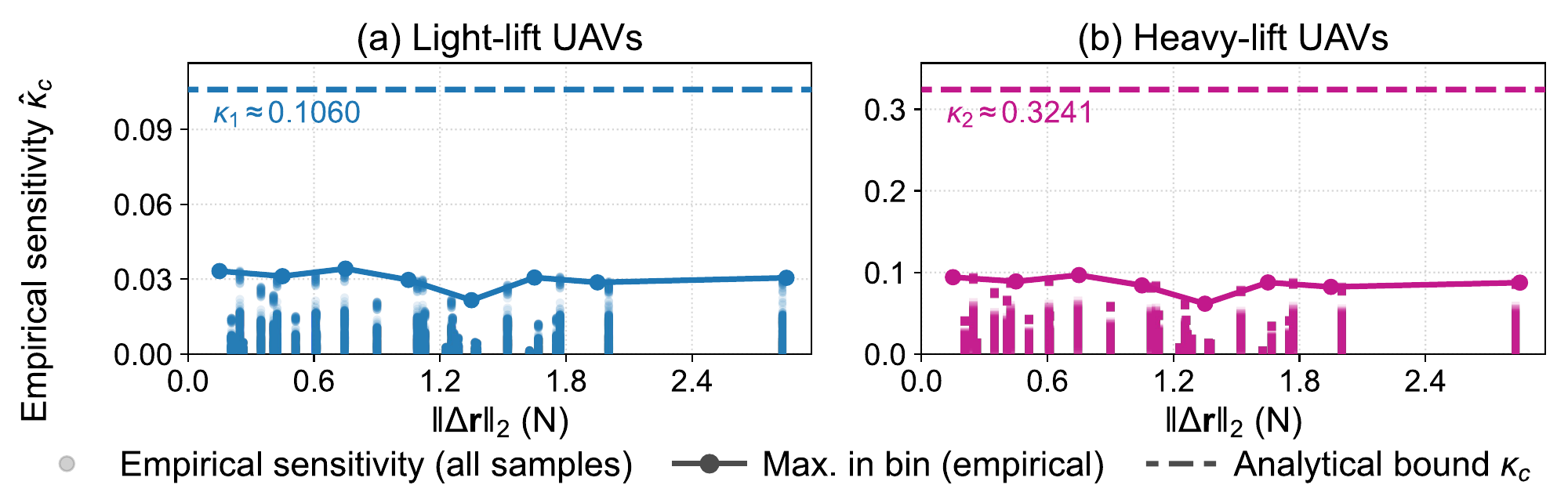}

\caption{Empirical validation of the analytically derived score sensitivity coefficients for (a) light-lift and (b) heavy-lift UAVs. Each point represents the change in mission-level score between two robustness margin vectors under the same mission realization $\omega$, normalized by the distance between the margin vectors. The dashed lines indicate the analytical bounds $\kappa_1=0.1060$ and $\kappa_2=0.3241$, respectively, and all sensitivities are below these bounds.}
    \label{fig:kappa}
    \vspace{-10pt}
\end{figure}

{The comparison times are $t_k=kh$, with $d_i^k=d_{i,k}$ in~\eqref{eq:incremental_sensitivity}.} The force allocation iterates are reinitialized at each sampling time. Following Prop.~\ref{prop:robot_kappa}, let $G_{0,j}=0$, $G_{k+1,j}=A_{k,j}G_{k,j}+b_{k,j}$, and $\bar G_j$ denote the componentwise maximum of $G_{k,j}$ for $k\in\{0,\ldots,100\}$, including the final comparison time after the $100$ scored intervals. The robot-level score coefficients in~\eqref{eq:score_sensitivity} are
\eqn{
\begin{aligned}
L_{i,j}&:=\left[(H_uC+H_x)_{i,:}\right]^\top,\qquad
L_{i,j}^{r}:=[H_ug]_i,\\
\kappa_{i,j}&:=L_{i,j}^{r}+\left(L_{i,j}\right)^\top\bar G_j,\qquad
\kappa_{c,j}:=\max_{i\in\Vcal_c}\kappa_{i,j}.
\end{aligned}
\label{eq:sim_payload_kappa}
}

{\Cref{fig:kappa} compares the coefficients from~\eqref{eq:sim_payload_kappa} with paired mission score sensitivities. The bounds use the controller parameters and force residual bounds, without requiring the residual distributions.}
We perform five \ac{MAICP} iterations with $M_j=500$ conditionally i.i.d. missions per iteration, $\alpha_c=0.1$, $\delta_{c,j}^{\mathrm{cal}}=0.05$, and $\ell_c=0$ for both classes. Since $K_c=3$, the mission-level score is the maximum robot-level score, which can be computed by flooding. We use $K_S=0$ and $K_q=100$ distributed pinball loss iterations followed by the finite-iteration upper bound computation. The initial margins are $(1.5,1.5)\,\mathrm N$, with $w_1=w_2=\lambda=1$ in~\eqref{eq:margin_update}. Under Assumption~\ref{ass:mission_sampling} and the feasibility condition~\eqref{eq:general_margin_constraint}, Theorem~\ref{th:maicp_coverage} provides calibration-conditional mission-level score coverage of at least $0.9$ with confidence at least $0.95$ for each class. We compare \ac{MAICP} with the non-robust allocation obtained by setting $\br=0$. The results in~\Cref{tab:payload_results} show empirical mission-level score coverage above $90\%$ for both classes. 
\begin{table}[h!]
    \centering
    \caption{Cooperative payload allocation results at the final iteration}
    \label{tab:payload_results}
    \small
    \begin{tabular}{@{}lcc@{}}
        \toprule
        \textbf{Metric} & \textbf{Non-robust}& \textbf{\ac{MAICP}} \\
        \midrule
        Margins $(r_1,r_2)$ (N) & $(0,0)$ & $(0.159,1.481)$ \\
        Mission coverage (\%) & $(0,0)$ & $(99.93,96.11)$ \\
        Mean max. $|e_i|$ (m) & 0.1053 & 0.0769 \\
        Mean robot RMS (m) & 0.0527 & 0.0557 \\
        \bottomrule
    \end{tabular}
    \vspace{-5pt}
\end{table}

\subsection{Distributed Formation Control with Obstacle Avoidance}
\label{sec:sim_cbf}

\begin{figure}
    \centering
    \includegraphics[width=\columnwidth]{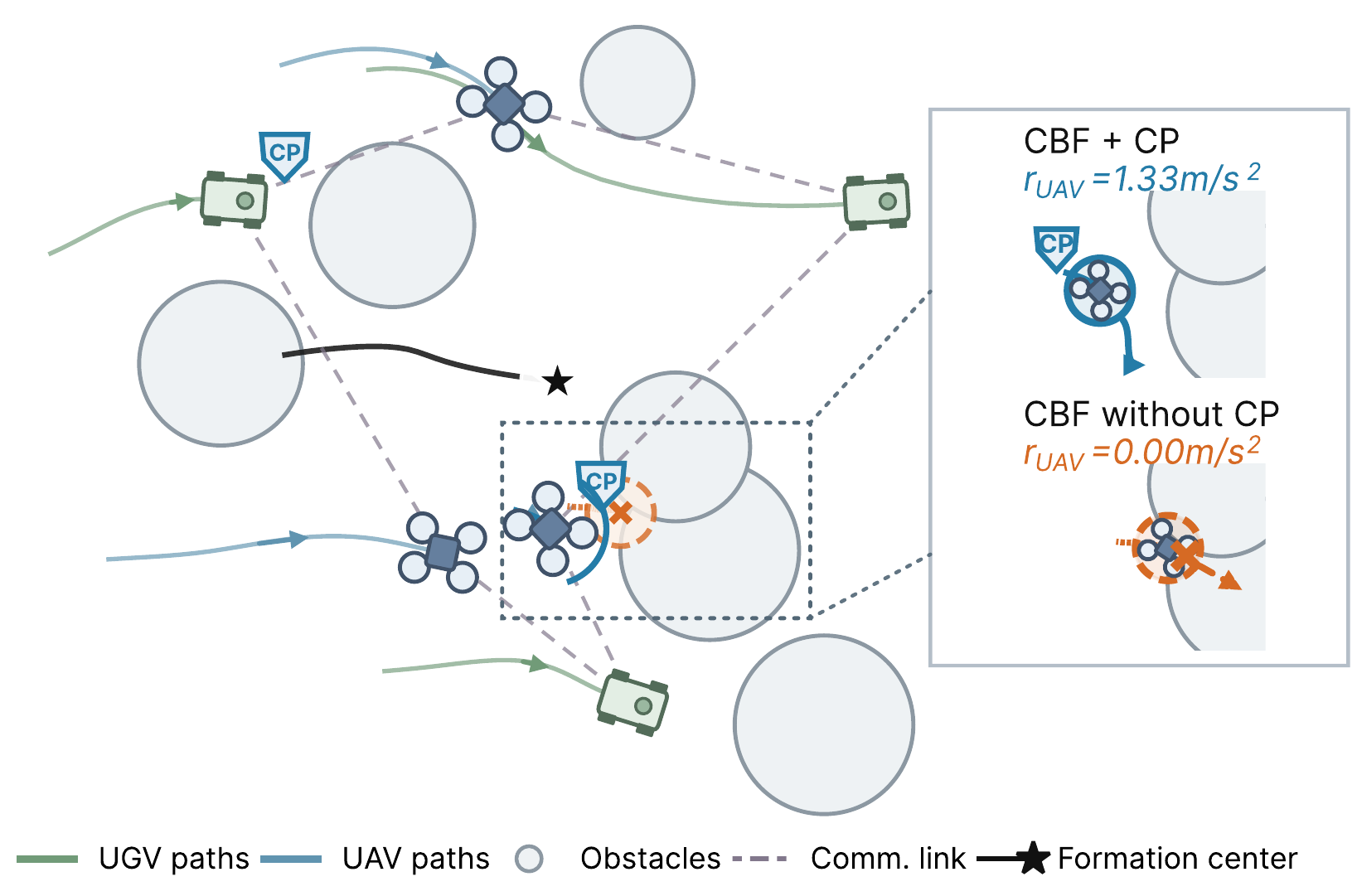}
    \caption{Formation control with three UGVs and three UAVs. The inset compares UAV obstacle avoidance using a CBF with robustness margin $r_{\mathrm{UAV}}=1.33\,\mathrm{m/s^2}$ and without robustness margins.}
    \label{fig:collision_environment}
    \vspace{-10pt}
\end{figure}

\begin{figure*}
    \centering
    \includegraphics[width=0.95\linewidth]{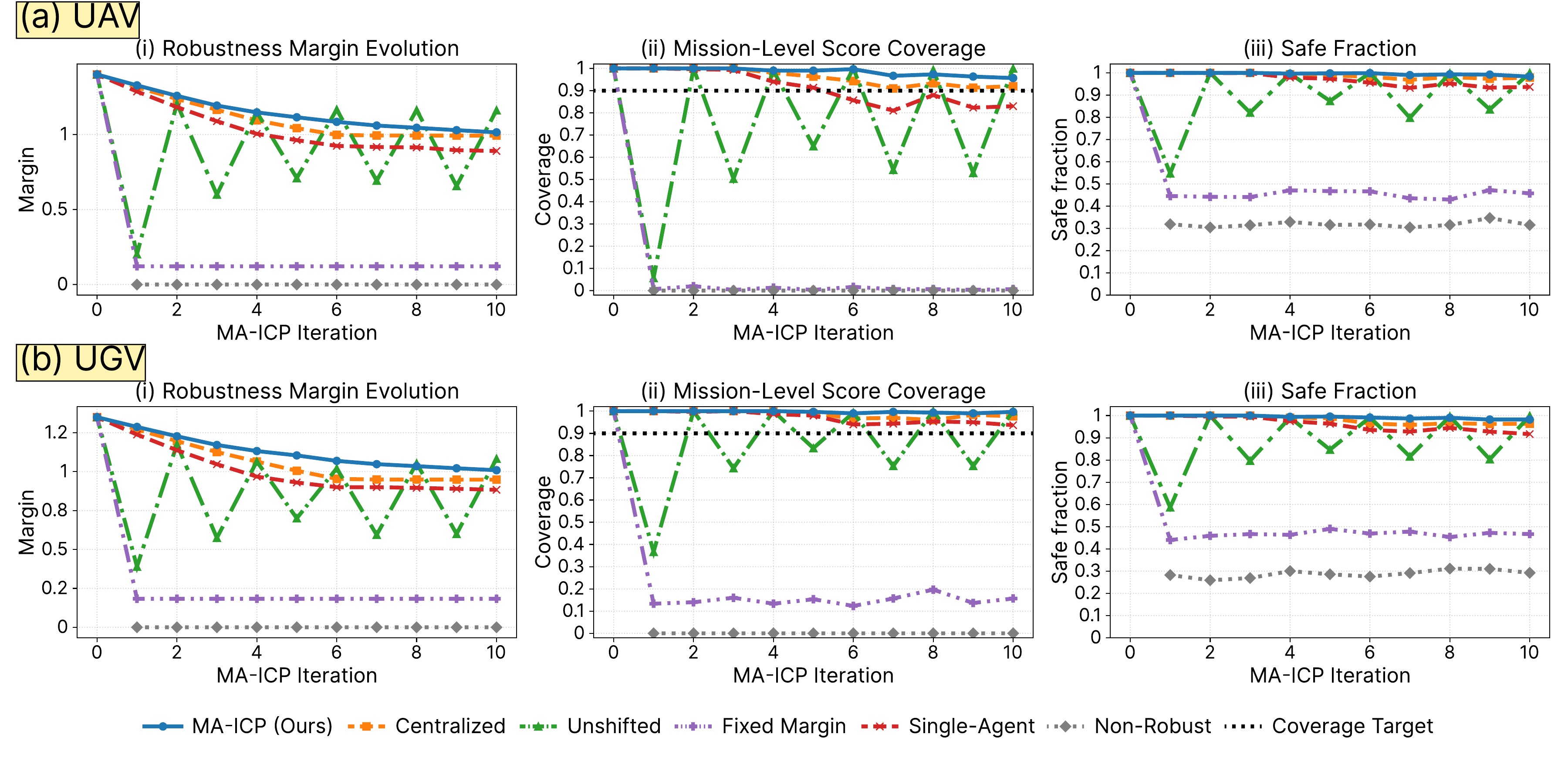}
    \caption{Formation control results for (a) UAVs and (b) UGVs. (i) Class-dependent robustness margins over \ac{MAICP} iterations. (ii) Mission-level score coverage, with the dashed line denoting the target coverage $1-\alpha_c=0.9$. (iii) Fraction of robots that avoid obstacle collisions throughout the mission.}
    \label{fig:collision_results}
    \vspace{-10pt}
\end{figure*}
{We consider distributed formation control with obstacle avoidance using $24$ robots in four independent subteams, each containing three unmanned ground vehicles (UGVs) and three UAVs.} UGVs occupy the ground layer, while UAVs operate at a fixed altitude. Each subteam maintains a formation with six equally spaced slots on a circle of radius $4\,\mathrm m$, alternating between UGVs and UAVs, while navigating around six cylindrical obstacles. The formation orientation and obstacle positions are randomized between missions. Each subteam executes $75$ independent missions per calibration iteration using the same class models in local subteam coordinates, contributing one mission-level score per class and mission. \Cref{fig:collision_environment} illustrates the formation and obstacle avoidance task.

For robot $i\in\Vcal$, let $p_i,v_i,u_i\in\R^2$ denote its position, velocity, and acceleration input, respectively, and let $\delta_i\in\R^2$ denote its fixed formation offset. The robot dynamics are
\eqn{
\dot p_i=v_i,
\dot v_i=u_i+d_{C(i)}^{\mathrm{pos}}(p_i)+d_i^{\mathrm{nbr}}(p)+d_i^{\mathrm{enc}}(p)+w_i.
\label{eq:sim_formation_dynamics}
}
{The map $d_{C(i)}^{\mathrm{pos}}:\R^2\to\R^2$ describes the position-dependent acceleration residual, while $d_i^{\mathrm{nbr}}$ and $d_i^{\mathrm{enc}}$ describe neighbor-induced and obstacle encounter acceleration residuals, respectively.} {The controller uses frozen learned approximations $\hat d_{C(i)}^{\mathrm{pos}}$ and $\hat d_i^{\mathrm{nbr}}$, trained offline.} The neighbor dynamics depend on the relative positions of both UGVs and UAVs. {Each mission scales the difference between the learned and true position and neighbor dynamics by an independently sampled factor distributed uniformly over $[0,2]$.} The obstacle encounter disturbance depends on obstacle proximity and the positions of robots from the other class. The process noise $w_i\in\R^2$ has standard deviation $0.01\,\mathrm{m/s^2}$ and is truncated such that $\norm{w_i}_2\leq0.03\,\mathrm{m/s^2}$. All of these exogenous disturbances are sampled independently and are independent from the robustness margins.

The robots estimate the formation center using a fixed communication ring that alternates between UGV and UAV robots. Robot $i$ forms the local center observation $y_{i,k}:=p_{i,k}-\delta_i$ and maintains the estimate $\hat c_{i,k}\in\R^2$. With one synchronous communication round per sample time and self and neighbor weights equal to $\frac{1}{3}$, the dynamic average consensus update is
\eqn{
\begin{aligned}
\hat c_{i,k+1}
={}
\sum_{h\in\Ncal_i\cup\{i\}}
w_{ih}\hat c_{h,k}
+y_{i,k+1}-y_{i,k}.
\end{aligned}
\label{eq:sim_center_consensus}
}
The estimates are initialized by $\hat c_{i,0}=y_{i,0}$. Since the mixing matrix is doubly stochastic, the average of the estimates equals the average of the center observations at each step.

The prescribed center reference $c_k^{\mathrm{ref}}\in\R^2$ follows a minimum jerk trajectory between the initial and terminal formation centers, with velocity $v_k^{\mathrm{ref}}$ and acceleration $a_k^{\mathrm{ref}}$. Robot $i$ computes the nominal formation control input
\eqn{
\begin{aligned}
u_{i,k}^{\mathrm{nom}}
={}a_k^{\mathrm{ref}}&
+k_{C(i)}^{\mathrm{form}}
(\hat c_{i,k}+\delta_i-p_{i,k})\\
&+k_{C(i)}^{\mathrm{goal}}
(c_k^{\mathrm{ref}}-\hat c_{i,k})-k_{C(i)}^{\mathrm{vel}}
(v_{i,k}-v_k^{\mathrm{ref}}).
\end{aligned}
\label{eq:sim_formation_nominal_control}
}
We use $(k_c^{\mathrm{form}},k_c^{\mathrm{goal}},k_c^{\mathrm{vel}})=(0.8,0.4,1.2)$ for UGVs and $(1.0,0.5,1.5)$ for UAVs. The formation and goal gains are distinct so that the center estimate remains in the feedback law. Changes in one robot trajectory therefore change the center observations used by the other robots through~\eqref{eq:sim_center_consensus}.


{Let $\hat d_i:=\hat d_{C(i)}^{\mathrm{pos}}(p_i)+\hat d_i^{\mathrm{nbr}}(p)$ and $d_i:=d_{C(i)}^{\mathrm{pos}}(p_i)+d_i^{\mathrm{nbr}}(p)+d_i^{\mathrm{enc}}(p)+w_i$ denote the learned and true acceleration residuals. Taking $z_i:=\operatorname{col}(\hat c_i,\hat d_i)$ and $z_i^\star:=\operatorname{col}(\bar y,d_i)$, where $\bar y$ is the mean of the six center observations, we measure only the acceleration error in Definition~\ref{def:robot_score}, i.e., $\rho_{C(i)}(z_i,z_i^\star)=\norm{d_i-\hat d_i}_2$. The robot-level score is $S_i(\omega;\br):=\sup_{t\in\Tcal_i^{\mathrm{score}}(\omega;\br)}\norm{d_i(t)-\hat d_i(t)}_2$.}
{Thus, $S_i(\omega;\br)\leq r_{C(i)}$ bounds the acceleration model error over the score evaluation times. Since the robustness margins modify the control inputs and robot trajectories, they also affect the residuals and the formation center estimates.}

{Each robot uses a quadratic program (QP) with robust high-order control barrier function (CBF) constraints~\cite{xiao2021high} for obstacle avoidance.} For a circular obstacle with center $p_o\in\R^2$, define $q_{io}:=p_i-p_o$ and $h_{io}:=\norm{q_{io}}_2^2-D_{io}^2$, where $D_{io}>0$ is the sum of the robot and obstacle radii. With $\lambda_1,\lambda_2\in\Rplus$, let $\psi_{io}:=2q_{io}^\top v_i+\lambda_1h_{io}$. For each obstacle, robot $i$ optimizes the coefficient $\mu_{io}\in[0,10]$ from~\cite{zeng2021safety} and imposes the robust constraint
\eqn{
\begin{aligned}
2q_{io}^\top(u_i+\hat d_i)
\geq{}&-2\norm{v_i}_2^2-2\lambda_1q_{io}^\top v_i\\
&-\mu_{io}\lambda_2\psi_{io}
+2\norm{q_{io}}_2r_{C(i)}.
\end{aligned}
\label{eq:sim_robust_cbf}
}
The final term bounds the effect of the acceleration model error on the barrier derivative whenever $S_i(\omega;\br)\leq r_{C(i)}$. The local QP jointly optimizes the acceleration input and obstacle decay multipliers, minimizing the deviation from $u_i^{\mathrm{nom}}$ with a quadratic penalty on deviations of $\mu_{io}$ from unity. We use $\lambda_1=\lambda_2=1.5$, robot radius $0.45\,\mathrm{m}$, and componentwise acceleration limits of $2\,\mathrm{m/s^2}$ and $4\,\mathrm{m/s^2}$ for UGVs and UAVs, respectively. {The formation case study uses a sampling period of $0.2\,\mathrm s$ and a maximum of 250 simulation steps per mission.}

{Both the formation and target tracking studies use ten \ac{MAICP} updates with $M_j=300$ missions per iteration, $\alpha_c=\delta_{c,j}^{\mathrm{cal}}=0.1$, $k_{c,j}=278$, and $K_S=K_q=100$. We set $D=I_2$, $w_c=1$, $\lambda=0.25$, and $\gamma_\ell=0.05/(\ell+1)^{0.75}$, with Metropolis weights $w_{ih}=[1+\max\{\deg(i),\deg(h)\}]^{-1}$ on the connected update graph and $w_{ii}=1-\sum_{h\neq i}w_{ih}$, where $\deg(i)$ is the number of neighbors of robot $i$. For formation control, $\ell_{\mathrm{UGV}}=1$ and $\ell_{\mathrm{UAV}}=0$, so at least two UGVs and all three UAVs must satisfy their score bounds. The initial margins are $1.35$ and $1.40\,\mathrm{m/s^2}$, respectively, within $[0,4]\,\mathrm{m/s^2}$. We use $\kappa_{\mathrm{UGV},j}=9.4$ and $\kappa_{\mathrm{UAV},j}=10.3$.}


We compare \ac{MAICP} with five variants. The \emph{centralized} method uses exact order statistics, while the \emph{unshifted} method sets $\kappa_{c,j}=0$. The \emph{fixed margin} variant keeps the margins computed from the first calibration iteration, and the \emph{non-robust} method sets $\br=0$. The \emph{single-agent} interactive CP variant~\cite{mirzaeedodangeh2026robust} independently updates each robot's margin using its own scores, the class miscoverage and calibration confidence parameters, and a configured scalar sensitivity coefficient, without accounting for changes in the margins of other robots. 
All methods begin calibration under the same initial robustness margins and use the same mission randomizations. The fixed margin variant displays the margins obtained from the first calibration iteration, which are then held constant, while the iterative methods use the scores collected at iteration $j$ to compute the margins deployed at iteration $j+1$.


Figure~\ref{fig:collision_results} shows that \ac{MAICP} reduces the robustness margins while maintaining empirical mission-level score coverage above $0.9$ for both classes. The centralized variant obtains slightly smaller margins using exact order statistics. The unshifted variant falls below the target coverage and has a lower fraction of collision-free robots after the first margin update, while the fixed margin and non-robust variants perform substantially worse. The single-agent variant also falls below the target mission-level coverage for UAVs after the fifth iteration, illustrating why the calibration must account for the coupling between robots.

\subsection{Distributed Target Tracking}
\label{sec:sim_tracking}

\begin{figure}
    \centering
    \includegraphics[width=\linewidth]{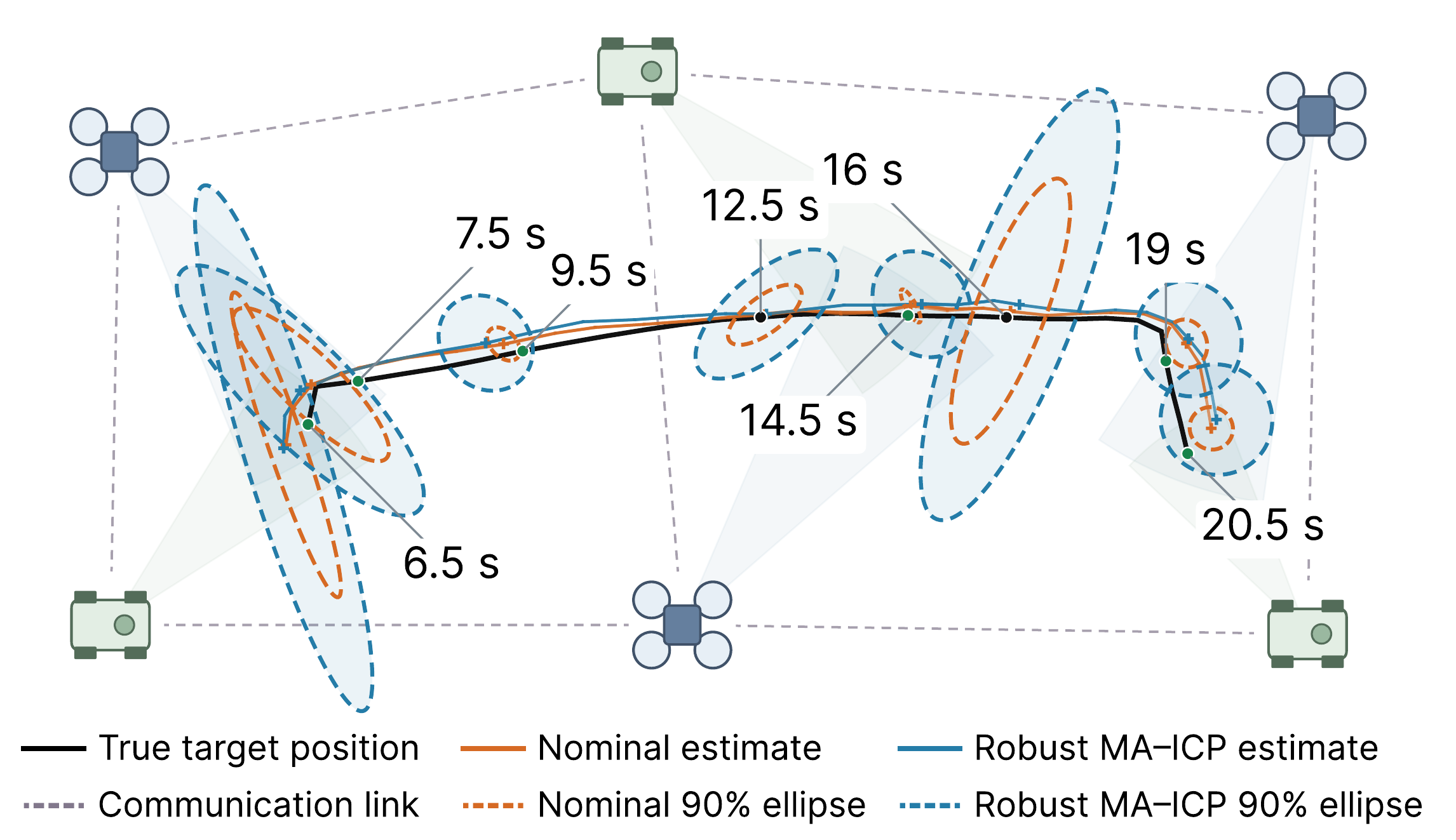}
    \caption{Representative target tracking mission with UAVs and UGVs. The nominal covariance ellipses frequently exclude the true target position, while the regions calibrated by \ac{MAICP} cover the target along the trajectory.}

    \label{fig:tracking_environment}
\end{figure}

\begin{figure*}
    \centering
    \includegraphics[width=0.95\textwidth]{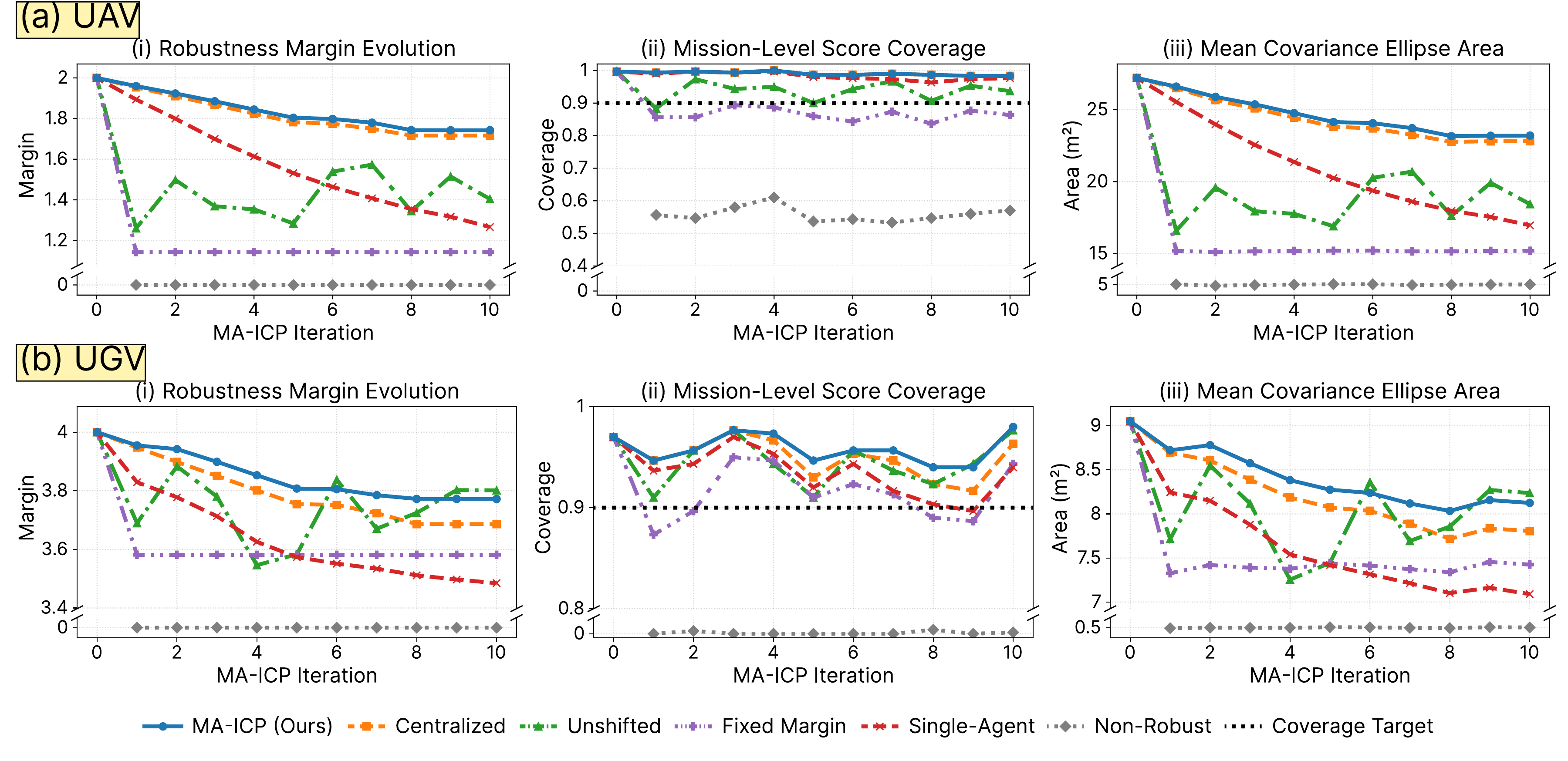}
    \caption{Distributed target tracking results for (a) UAVs and (b) UGVs. (i) Class-dependent robustness margins over \ac{MAICP} iterations. (ii) Mission-level score coverage, with the dashed line denoting the target coverage $1-\alpha_c=0.9$. (iii) Mean area of the target position covariance ellipses.}
    \label{fig:tracking_results}
    \vspace{-10pt}
\end{figure*}

We consider distributed target tracking for heterogeneous UGV and UAV teams using the Distributed Rolling Window Tracking (DRWT) algorithm from~\cite{shorinwa2020distributed}. DRWT solves a maximum \textit{a posteriori} estimation problem across a rolling window in a distributed manner using the alternating direction method of multipliers (ADMM). Let $x_k^{\mathrm{tar}}=[(p_k^{\mathrm{tar}})^\top,(v_k^{\mathrm{tar}})^\top]^\top\in\R^4$ denote the planar target state. The estimator uses the constant-velocity model $x_{k+1}^{\mathrm{tar}}=Ax_k^{\mathrm{tar}}+w_k$, where $A=\left[\begin{smallmatrix}1&\Delta t\\0&1\end{smallmatrix}\right]\otimes I_2$, $\Delta t=0.5\,\mathrm{s}$, and $\otimes$ denotes the Kronecker product. $w_k$ is the process noise with covariance $Q=10^{-4}I_4+0.04^2(hh^\top\otimes I_2)$, where $h:=(\Delta t,1)^\top$.
The simulated target's velocity is incremented with $\eta_k\sim\mathcal N(0,0.04^2I_2)$ and is saturated to a magnitude of $2.2\,\mathrm{m/s}$. The target moves within a bounded region in a zig-zag pattern. The initial target position and velocity are randomized with standard deviations $0.45\,\mathrm{m}$ and $0.08\,\mathrm{m/s}$, respectively. The tracker is initialized with covariance $P_0=\operatorname{diag}(0.45^2I_2,0.08^2I_2)$.

{Four independent subteams, each containing five UGVs and five UAVs, track separate targets and contribute $75$ missions each per calibration iteration. Robots follow fixed waypoint routes independently of the target motion. The UGVs use unicycle dynamics with lateral slip and drag coefficients $0.045$ and $0.018$, respectively, while the UAVs use planar double-integrator dynamics at $12\,\mathrm{m}$ altitude with drag coefficient $0.035$.  Their maximum commanded speeds are $5\,\mathrm{m/s}$ and $7\,\mathrm{m/s}$, respectively. These disturbances are included in $\omega$.}

The two robot classes use distinct sensing models. UGVs measure target range and bearing using planar LiDAR, while UAVs obtain bearing-only measurements with a monocular camera. For robot $i$, let $\rho_{i,k} \in \Rnonneg$ denote the target range, $\phi_{i,k}\in(-\pi,\pi]$ the true bearing relative to its heading $\psi_{i,k}\in(-\pi,\pi]$, and $p_{i,k}\in\R^2$ its planar position. \Cref{fig:tracking_environment} shows a representative tracking mission and the nominal and calibrated uncertainty ellipses.


{The UGV measures range $\widetilde\rho_{i,k}=\rho_{i,k}+b_i^\rho+\nu_{i,k}^\rho$ and bearing $\widetilde\phi_{i,k}=\phi_{i,k}+b_i^\phi+\nu_{i,k}^\phi$, yielding the planar observation $y_{i,k}=p_{i,k}+\widetilde\rho_{i,k}(\cos(\psi_{i,k}+\widetilde\phi_{i,k}),\sin(\psi_{i,k}+\widetilde\phi_{i,k}))^\top$.} 
The LiDAR has a maximum range of $25\,\mathrm{m}$ and a $270^\circ$ field of view. Its range and bearing noise standard deviations are $0.20\lambda_m\,\mathrm{m}$ and $0.020\lambda_m\,\mathrm{rad}$, respectively, where $\lambda_m\sim\operatorname{Unif}[0.7,1.3]$ is sampled independently for each mission. {Each UGV also has persistent range and bearing biases sampled uniformly from $[-0.15,0.15]\,\mathrm{m}$ and $[-0.010,0.010]\,\mathrm{rad}$. The nominal measurement covariance is $R_{i,k}^{\mathrm{sens}}=J_{i,k}\operatorname{diag}((0.20\lambda_m)^2,(0.020\lambda_m)^2)J_{i,k}^\top$, where $J_{i,k}$ is the Jacobian of $y_{i,k}$ with respect to $(\widetilde\rho_{i,k},\widetilde\phi_{i,k})$.}

For a UAV, the monocular camera provides a noisy bearing $\widetilde\phi_{i,k}=\phi_{i,k}+b_i^{\mathrm{cam}}+\nu_{i,k}^{\mathrm{cam}}$, where $\nu_{i,k}^{\mathrm{cam}}\sim\mathcal N(0,0.025^2)$ and $b_i^{\mathrm{cam}}\sim\operatorname{Unif}[-0.015,0.015]$ is fixed throughout the mission. The camera has a $120^\circ$ field of view and maximum detection range of $50\,\mathrm{m}$. 
{The measured bearing $\theta_{i,k}:=\psi_{i,k}+\widetilde\phi_{i,k}$ defines $n_{i,k}:=(-\sin\theta_{i,k},\cos\theta_{i,k})^\top$. The line-of-sight observation is $y_{i,k}=n_{i,k}^\top p_{i,k}$, with measurement matrix $H_{i,k}=[n_{i,k}^\top\ \ 0_{1\times2}]$. Thus, each camera observation is scalar and doesn't contain target range information.} The range-dependent measurement covariance is $R_{i,k}^{\mathrm{sens}}=(0.025\hat\rho_{i,k}^{\mathrm{prior}})^2+0.05^2$, where $\hat\rho_{i,k}^{\mathrm{prior}}$ is the target range computed from the robot's local estimate before the current DRWT update. 

For both classes, the robustness margin modifies the measurement covariance used by the distributed estimator as $R_{i,k}(r_{C(i)})=R_{i,k}^{\mathrm{sens}}+\beta_{C(i)}r_{C(i)}I_{m_i}$, where $m_i=2$ for UGVs and $m_i=1$ for UAVs, with $\beta_{\mathrm{UGV}}=0.02$ and $\beta_{\mathrm{UAV}}=0.05$. We use a four-step rolling window, $50$ ADMM iterations per estimator update, penalty parameter $0.8$, and a distance-based communication graph with radius $30\,\mathrm{m}$. Changing either class margin modifies its contribution to the distributed estimation problem and therefore affects the target estimates maintained by robots from both classes.

Let $\hat p_{i,t}(\br)$ and $P_{i,t}(\br)\succ0$ denote the estimated target position and its corresponding covariance. 
{After each mission, we use the recorded target position to evaluate the tracking error} $e_{i,k}:=\hat p_{i,k}(\br)-p_k^{\mathrm{tar}}$. 
{Define $q_{0,c}:=\sqrt{\chi^2_{2,1-\alpha_c}}$ as the nominal coverage ellipse radius, where $\chi^2_{2,1-\alpha_c}$ denotes the $(1-\alpha_c)$ quantile of a chi-squared distribution with two degrees of freedom. Let} 
{$\Tcal_i^{\mathrm{val}}(\omega;\br)$ denote the nonempty set of estimator update indices {for which the estimator returns a target estimate and positive definite covariance under $(\omega,\br)$}. The corresponding update times form $\Tcal_i^{\mathrm{score}}(\omega;\br)$ in Definition~\ref{def:robot_score}.} The robot-level score is $S_i(\omega;\br):=\left[\max_{k\in\Tcal_i^{\mathrm{val}}(\omega;\br)}\sqrt{e_{i,k}^\top P_{i,k}^{-1}e_{i,k}}-q_{0,C(i)}\right]_+$.

{The margin inflates the reported covariance to $P_{i,k}^{\mathrm{cal}}=((q_{0,C(i)}+r_{C(i)})/q_{0,C(i)})^2P_{i,k}$. The corresponding ellipse consists of positions $p\in\R^2$ satisfying}
\eqnN{
{(p-\hat p_{i,k})^\top P_{i,k}^{-1}(p-\hat p_{i,k})\leq(q_{0,C(i)}+r_{C(i)})^2.}
}
{Thus, $S_i(\omega;\br)\leq r_{C(i)}$ is equivalent to the target lying within this ellipse at every valid tracker update. When $r_c=0$, this reduces to the nominal ellipse; positive margins affect both the estimator through $R_{i,k}(r_c)$ and the reported uncertainty.}
{With the shared calibration settings given in~\Cref{sec:sim_cbf}, both classes use $\ell_c=1$, so four of the five robots must satisfy their score bounds. The initial margins are $r_{\mathrm{UGV},0}=4$ and $r_{\mathrm{UAV},0}=2$ within $[0,100]$. We use $\kappa_{\mathrm{UGV},j}=1.6$ and $\kappa_{\mathrm{UAV},j}=11.6$.}

\begin{figure*}[!t]
    \centering
    \includegraphics[width=0.95\textwidth]{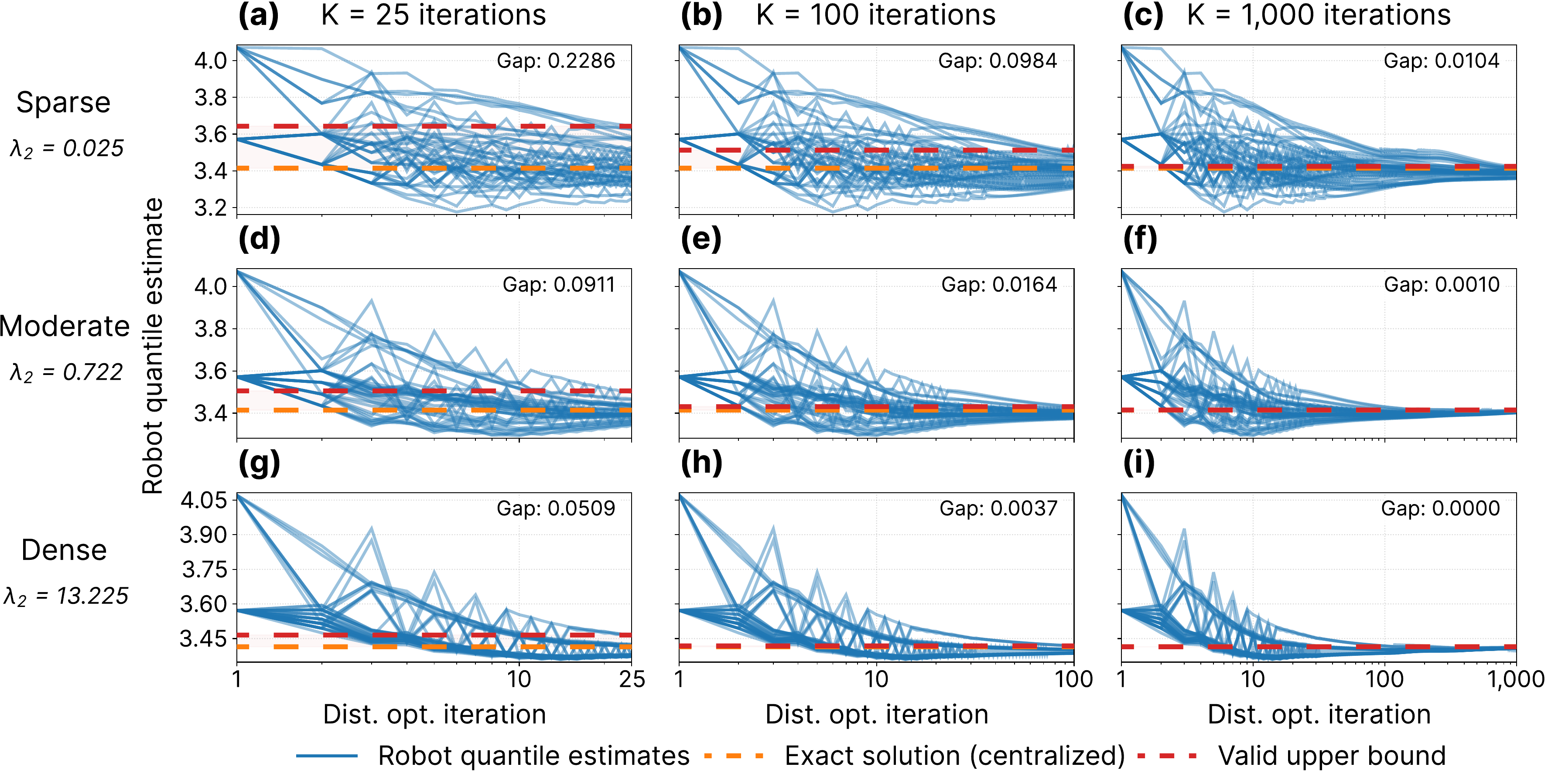}
\caption{Distributed computation of the quantile using 200 mission-level scores from the target tracking experiment. Forty robots collaboratively compute the 182nd-smallest score, with each robot holding five scores. Rows correspond to communication graphs with 2, 8, and 24 neighbors per robot, respectively, and columns show $K=25$, $100$, and $1000$ distributed subgradient iterations. The blue curves show the local estimates $q_i^\ell$, the orange dashed lines indicate the exact centralized order statistic $a_{[182]}=3.414451$, and the red dashed lines indicate the upper bounds $\bar a_{[182]}^K$ computed after $K$ iterations.}    
\label{fig:quantile}
\vspace{-10pt}
\end{figure*}

{We compare the same five variants described in~\Cref{sec:sim_cbf}. The non-robust tracker uses zero robustness margin and reports its nominal uncalibrated ellipse. For the single-agent variant, the plotted class margin is the mean of the individual robot margins, while mission-level coverage is evaluated using each participating robot's own margin.}

Figure~\ref{fig:tracking_results} shows that \ac{MAICP} reduces the robustness margins and covariance ellipse areas while maintaining empirical mission-level coverage above $0.9$ for both classes. The centralized variant obtains slightly smaller margins using exact order statistics. The unshifted variant does not consistently maintain the target coverage, while the fixed margin and non-robust variants have substantially lower coverage. The single-agent variant produces smaller uncertainty ellipses but briefly falls below the target coverage for UGVs.

\subsection{Finite-Iteration Order Statistic Upper Bound}
\label{app:order_statistic}

To illustrate the finite-iteration guarantee in Proposition~\ref{prop:finite_order_statistic}, we consider the distributed quantile computation using data collected from the target tracking experiment in~\Cref{sec:sim_tracking}. We use a fixed dataset of $M=200$ mission-level scores for the UGV class, where each mission-level score is the fourth-smallest of the five participating UGV robot-level scores. The mission-level scores are computed exactly so that the experiment isolates the distributed quantile computation. We consider the order statistic with rank $k=182$, whose exact centralized value is $a_{[182]}=3.414451$. The 200 scores are distributed across 40 robots, consisting of four subteams with five UGVs and five UAVs each, with every robot holding five scores. {The assignment of five scores per robot is a disjoint partition used to isolate the distributed quantile computation and is distinct from the mission participation sets.}

The robots compute a common estimate of this order statistic using~\eqref{eq:distributed_quantile_update} with step size $\gamma_\ell=0.5/(\ell+1)^{0.75}$. {We consider three connected communication graphs with 2, 8, and 24 neighbors per robot, with algebraic connectivity values $\lambda_2=0.025$, $0.722$, and $13.225$ and graph diameters $D_j=20$, $5$, and $2$, respectively, where $\lambda_2$ is the second-smallest eigenvalue of the graph Laplacian.} The graphs use the same robots and calibration data, and differ only in their communication topology. For each graph, we perform $K\in\{25,100,1000\}$ distributed subgradient iterations and compute the corresponding upper bound $\bar a_{[182]}^K$ using Algorithm~\ref{alg:distributed_order_statistic}.

As shown in~\Cref{fig:quantile}, the local estimates $q_i^\ell$ do not necessarily upper bound the exact order statistic after a finite number of iterations. However, {the computed upper bounds satisfy} $a_{[182]}\leq\bar a_{[182]}^K$ in all cases, consistent with Proposition~\ref{prop:finite_order_statistic}. {In this experiment, increasing the number of iterations generally reduces the gap between the computed upper bound and the exact order statistic}, while the more connected graphs exhibit faster convergence. Thus, {the upper bound computation is valid without convergence of the distributed optimization}.

\section{Conclusion}

In this paper, we introduced \ac{MAICP}, a distributed framework for computing robustness margins in heterogeneous multi-robot systems in which the calibrated margins induce distribution shifts through the coupled distributed algorithms and closed-loop behavior of the robots. By aggregating robot-level scores into mission-level scores, our method allows dependence among robots within a mission and combines finite-iteration distributed quantile computations with score sensitivity bounds to establish calibration-conditional coverage guarantees for each robot class. We demonstrated the method on cooperative payload force allocation, distributed collision avoidance, and target tracking.

Potential applications beyond those considered here include cooperative localization, distributed mapping, and other multi-robot estimation and control problems. Future work will investigate more data-efficient calibration methods, including reusing data collected under previous robustness margins and developing anytime-valid coverage guarantees that allow the number of calibration missions and the timing of margin updates to be selected adaptively.

\bibliographystyle{IEEEtran}
\bibliography{biblio, IEEEabrv}
\begin{authorbiography}
[{\includegraphics[width=1in,height=1.25in,clip,keepaspectratio]
{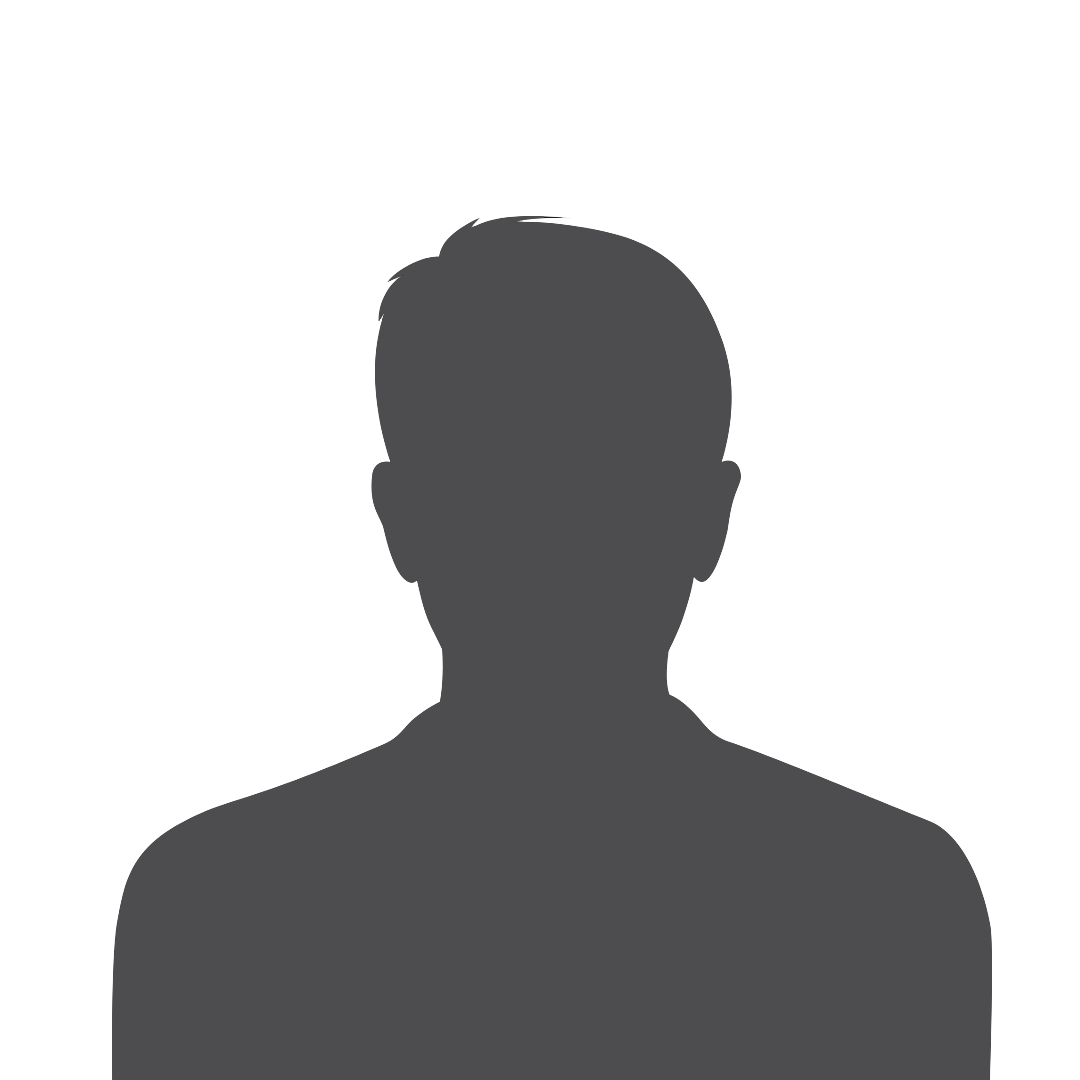}}]
{Daniel Cherenson}is a Ph.D. candidate in Robotics at the University of Michigan. He received the M.S. and B.S. degrees in electrical engineering and mechanical engineering, respectively, from Washington University in St. Louis. His research interests include safety-critical planning and control under uncertainty, with emphasis on robust safety, uncertainty quantification, and online adaptation.
\end{authorbiography}

\begin{authorbiography}[{\includegraphics[width=1in,height=1.25in,clip,keepaspectratio]{figs/Headshot-Placeholder.png}}]{Dimitra Panagou}received the Diploma and Ph.D. degrees in mechanical engineering from the National Technical University of Athens. She is currently a Full Professor with the Department of Robotics and the Department of Aerospace Engineering at the University of Michigan. Her research interests include nonlinear systems and control, multiagent systems, autonomy, and aerospace robotics, with emphasis on provably correct methods for the safe and resilient operation of autonomous systems under uncertainty. She is a recipient of the NASA Early Career Faculty Award, the AFOSR Young Investigator Award, and the NSF CAREER Award.
\end{authorbiography}
\end{document}